\documentclass[]{artificer}
\usepackage{microtype}
\usepackage{amsfonts}
\usepackage{graphicx}
\usepackage{booktabs}
\usepackage{wrapfig}
\usepackage{amsmath}
\usepackage{amsthm}
\usepackage{marvosym}
\usepackage{algpseudocode}

\usepackage[linesnumbered,lined,boxed,commentsnumbered,ruled,longend]{algorithm2e}
\usepackage[capitalize,noabbrev]{cleveref}
\usepackage{enumitem}

\usepackage{tcolorbox}
\usepackage{xcolor}
\colorlet{tabcolor}{colorbg}

\newcommand{\bx}{\mathbf x}

\newcommand{\bc}{\mathbf c}

\newcommand{\bs}{\mathbf s}

\newcommand{\bF}{\mathbf F}

\newcommand{\bW}{\mathbf W}
\newcommand{\bA}{\mathbf A}
\newcommand{\bI}{\mathbf I}

\newcommand{\bB}{\mathbf B}

\newcommand{\mcN}{\mathcal N}

\newcommand{\mcZ}{\mathcal Z}

\newcommand{\eg}{\textit{e.g.}}
\newcommand{\ie}{\textit{i.e.}}

\title{HyperAlign: Hypernetwork for Efficient Test-Time Alignment of Diffusion Models}

\author[1]{Xin Xie}
\author[2]{Jiaxian Guo}
\author[1 \, {\textnormal{\Letter}}]{Dong Gong}
\affiliation[1]{University of New South Wales (UNSW Sydney)}
\affiliation[2]{Google Research}
\emails{\{xin.xie3, dong.gong\}@unsw.edu.au, jeffguo@google.com}
\contribution{\textsuperscript{\Letter}Corresponding author.}

\abstract{
 Diffusion model alignment aims to bridge the gap between generated outputs and human preferences by enhancing both semantic consistency with textual prompts and overall visual quality. Existing alignment methods face a challenging trade-off: test-time approaches enable input-specific adaptability but introduce significant computational overhead and tend to under-optimize, while fine-tuning approaches risk reward over-optimization and loss of generation diversity. To bridge this gap, we propose HyperAlign, a framework that trains a hypernetwork for efficient and effective test-time alignment. Instead of modifying latent states directly, HyperAlign dynamically generates input-and-state-conditioned low-rank adaptation weights to modulate the denoising trajectory toward target rewards. We introduce multiple HyperAlign variants of varying granularity to balance alignment quality and computational efficiency. The hypernetwork is optimized with a reward objective regularized by preference data to mitigate reward hacking. We evaluate HyperAlign across multiple generative paradigms, including Stable Diffusion and FLUX, where it significantly outperforms existing alignment methods in semantic consistency and visual quality.
}

\metadata[Project Page]{\href{https://shelsin.github.io/hyperalign.github.io/}{hyperalign.github.io}}

\begin{document}

\maketitle

\vspace{0.8cm}
\begin{figure}[h]
\centering
\includegraphics[width=1.0\textwidth]{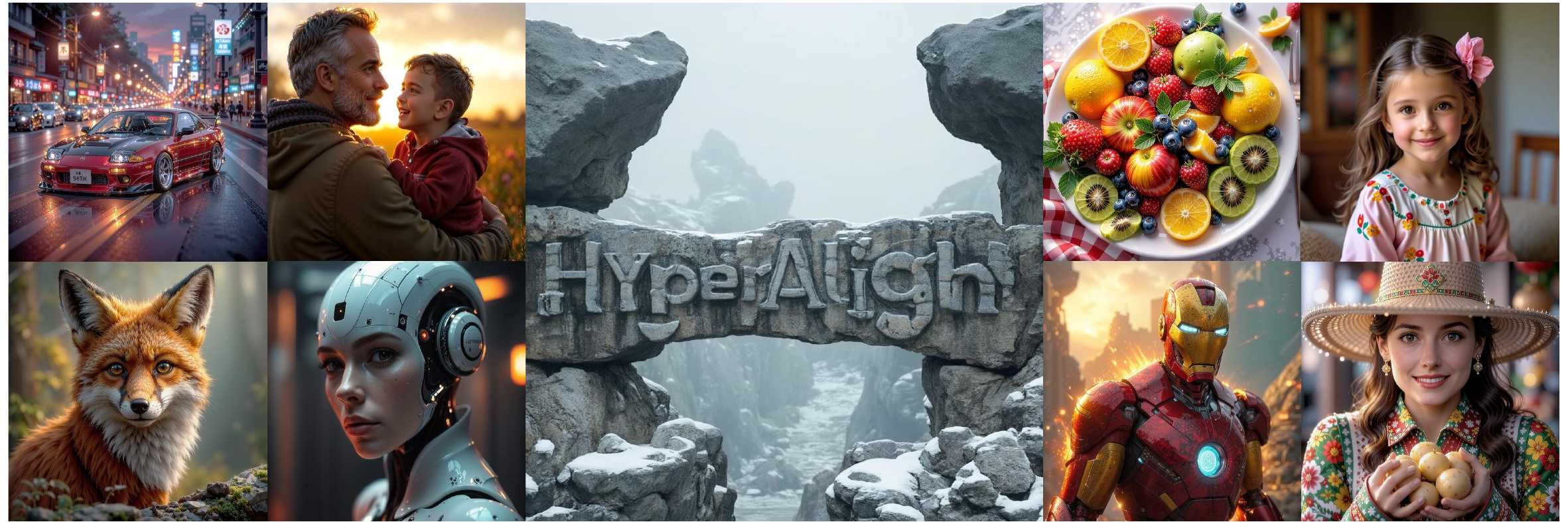} 
\caption{Sample images generated by our method based on FLUX backbone. The generated images not only achieve a high alignment with text prompt and human preferences, but also exhibit visually attractive and stunning aesthetics.}
\label{fig:first_image}
\end{figure}

\vspace{0.4cm}

\section{Introduction}
\label{sec:intro}

Diffusion models learn score function \cite{SDE} to gradually transform a random noise into a structured output, offering state-of-the-art performance in Text-to-Image (T2I) generation \cite{SD, SDXL, FLUX}. However, these models are typically trained on a large set of pre-collected datasets (\eg, web images) that may not accurately represent target conditional distribution aligned with human intention and preferences. 
As a result, generated images often fail to faithfully follow users' textual instructions or reflect their aesthetic preferences. While classifier-based and classifier-free guidance \cite{ClassifierGuidance, ClassifierFreeGuidance} improve prompt controllability, they do not address fine-grained human preferences. These challenges highlight the necessity of diffusion model \emph{alignment} \cite{Alignment}, which aims to bridge the gap between generated outputs and human preferences by enhancing both semantic consistency with textual prompts and overall visual quality.

\begin{wrapfigure}{r}{0.5\linewidth}
\vspace{-0.2cm}
\centering
\includegraphics[width=0.95\linewidth]{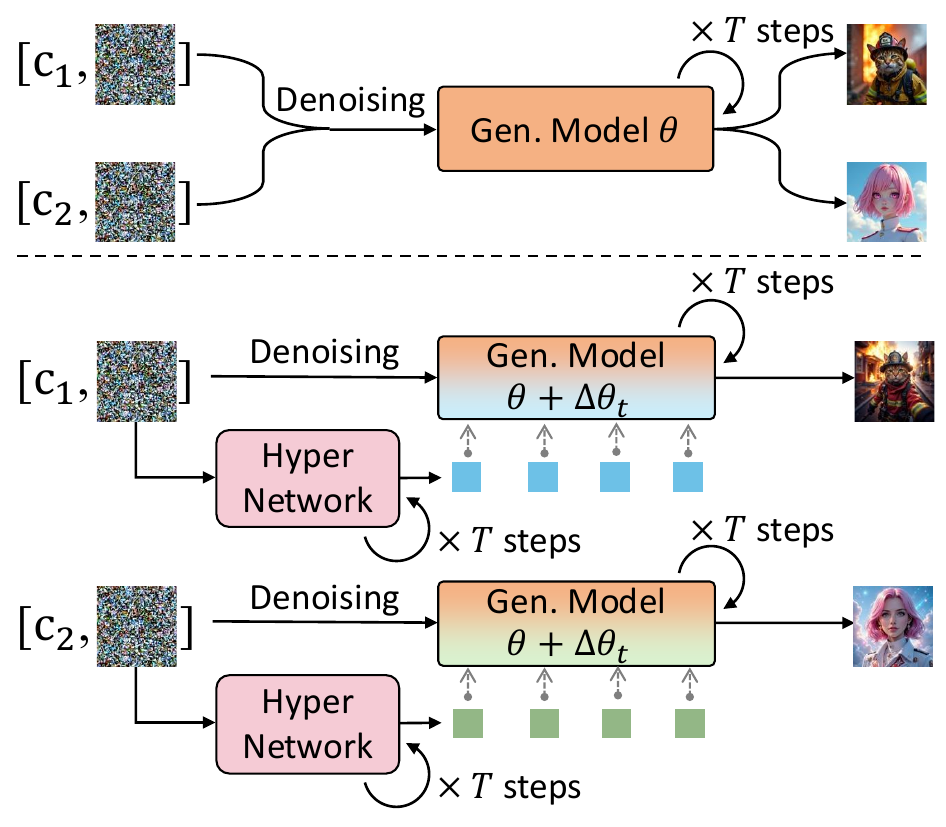}
\caption{Task-specific test-time alignment of HyperAlign. Compared to the original generative model, HyperAlign adapts the model’s behavior to each combination of prompt and temporal states, producing aligned and visually appealing results. }
\label{fig:iteration}
\vspace{-0.2cm}
\end{wrapfigure}

Diffusion model alignment can be approached through fine-tuning and test-time scaling. Fine-tuning alignment approaches, including Reinforcement Learning (RL) \cite{SPO, DiffusionDPO, DiffusionKTO, DanceGRPO, MixGRPO} and direct backpropagation \cite{SRPO, AlignProp, DRaFT, DRTune}, optimize target rewards based on explicit reward signals or implicit feedback from preference datasets \cite{PickAPic, HPS}. While training-based alignment methods effectively reshape the distribution to close to a desired target distribution, their generalization ability remains limited. Since inputs to the model, including textual prompts and sampled initial noise, vary considerably across use cases, a fixed set of fine-tuned parameters cannot sufficiently account for every combination of user intent, as illustrated in Fig.~\ref{fig:iteration}. Consequently, fine-tuning-based alignment often suffers from reward over-optimization (\eg, reward hacking) \cite{DAS, Alignment,DRaFT, AIG_Diversity}, leading to a loss of sample diversity and generation quality.

Test-time scaling methods, on the other hand, perform input-specific computing for the alignment goal during inference, through gradient-based approaches \cite{FreeDoM, DyMO, DAS, DNO} or sampling-based approaches \cite{BoN, EpsGreedy}. 
This allows the alignment process to be dynamically tailored to each individual query and intermediate denoising state, adjusting the denoising trajectory \cite{FreeDoM, DyMO, TFG} or the initial sampled noise \cite{ReNO, InitNO} to meet query-specific objectives. 
However, these methods incur substantial computational overhead from gradient calculations and repeated forward passes. Furthermore, they tend to suffer from reward under-optimization, as externally applied test-time guidance separated from the model's learned dynamics limits effective reward optimization. 

To bridge this gap, we propose \textbf{HyperAlign}, a hypernetwork-based framework that aims to combine the input-specific adaptability of test-time methods with the efficiency of training-based approaches. While test-time alignment methods produce input-specific results, they inevitably increase inference cost and react only to instantaneous rewards, limiting practical deployment. To retain input-specificity with improved efficiency, we learn a hypernetwork \cite{Hypernetworks, HyperNetField, Hyperdreambooth,text2lora} that generates input-and-state-conditioned adaptation weights on the fly, replacing expensive test-time computations. While the hypernetwork idea has been explored for amortizing test-time alignment \cite{HyperNoise} by predicting input-conditioned initial noise, it only focuses on distilled one-step models, \eg, \cite{FLUX_schnell,SDTurbo} (with post-hoc multi-step inference). In contrast, we explore and develop HyperAlign for general multi-step diffusion models. Leveraging the score-based nature of diffusion processes, we formulate the alignment objective as steering the denoising trajectory at each step toward a target reward. Instead of computing gradients at each step or fine-tuning the full parameter set, our hypernetwork dynamically generates input-and-step-conditioned low-rank adaptation weights (LoRA), which are applied on-the-fly to the frozen pre-trained model to guide intermediate denoising steps toward the reward. We further introduce three weight generation strategies of varying granularity: step-wise generation for fine-grained alignment, single generation at the starting point for minimal overhead, and piece-wise generation that updates adapters only at key timesteps to balance performance and efficiency. The hypernetwork is learned in post-training using reward supervision. Preference data is used to regularize the trajectories further mitigate reward hacking by preventing the model from overfitting to proxy scores while maintaining fidelity to genuine human preferences. We develop HyperAlign across two generative paradigms, diffusion models \cite{SD} and rectified flows \cite{FLUX}.

The main contributions are summarized as:

\begin{itemize}
    \item We propose HyperAlign, a hypernetwork-based framework for test-time alignment of diffusion models that enables the input-specific adaptability of test-time methods with efficiency and learned knowledge of trained models, applicable to general multi-step diffusion models and rectified flows.

    \item We achieve HyperAlign by training a hypernetwork that dynamically generates input-and-state-conditioned LoRA weights for pre-trained diffusion models, steering the denoising trajectory toward target rewards. Three weight generation strategies of varying granularity are introduced to balance alignment quality and efficiency. Input-specific adaptation combined with preference regularization mitigates reward hacking and preserves diversity.

    \item We validate HyperAlign on multiple generative architectures, including SD V1.5 and FLUX, demonstrating consistent improvements over state-of-the-art alignment methods across both quality and efficiency metrics.
\end{itemize}

\section{Related Work}
\label{sec:related_work}
\subsection{Fine-tuning Diffusion Model Alignment}
Diffusion models \cite{SD, SDXL, FLUX} exhibit remarkable generative performance, yet suffer from misalignment with human expectations. Early works \cite{AlignProp, ImageReward, DRaFT, DRTune, RAFT} directly learn preferences from reward models, but are constrained by unstable long-trajectory gradients. Therefore, SRPO \cite{SRPO} employs a noise prior to predict clear data and yields accurate reward gradient for each step. Alternatively, DDPO \cite{DDPO} and DPOK \cite{DPOK} integrate RL to optimize the score function \cite{SDE} through policy gradient updates. Subsequently, D3PO \cite{D3PO} and Diffusion-DPO \cite{DiffusionDPO} first introduce offline Direct Preference Optimization (DPO), modeling human preferences from win–lose paired data. SPO \cite{SPO} and LPO \cite{LPO} extend step-wise preference alignment by training timestep-aware reward models. Diffusion-KTO \cite{DiffusionKTO} adopt human utility maximization to reduce reliance on offline paired data. Recently, Flow-GRPO \cite{FlowGRPO} and DanceGRPO \cite{DanceGRPO} pioneer the integration of group-wise policy optimization paradigm \cite{GRPO} for improved diffusion model alignment. TempFlow-GRPO \cite{TempGRPO} designs temporal-aware credit assignment for intermediate-step advantage estimation and Pref-GRPO \cite{PrefGRPO} replaces pointwise score maximization objective with pairwise preference fitting. To improve the training efficiency, MixGRPO \cite{MixGRPO} adopts a mixed ODE–SDE paradigm and BranchGRPO \cite{BranchGRPO} restructures rollout process into a branching tree. Despite substantial advances in aligning diffusion models, discrepancies with human preferences and considerable computational burdens continue to pose challenges.

\subsection{Test-time Computing for Diffusion Models}
The goal of test-time scaling is to spend additional compute during inference to obtain more desirable generations. One naive scaling law in diffusion model sampling is to increase the number of denoising steps \cite{restart, zigzag}, enabling marginal improvements. Beyond this, there are two mainstream test-time techniques: One is sampling-based strategies, relying the reward models to evaluate multiple noise candidates and select more favorable denoising trajectory, such as Best-of-N search \cite{BoN}, evolutionary search \cite{evolutionary}, $\varepsilon$-greedy search \cite{EpsGreedy}, etc. The other one is gradient-based approaches, built on the score-based formulation of diffusion models \cite{SDE}. These methods use the differentiable reward functions to iteratively refine noise \cite{ReNO, InitNO, DNO}, prompt embeddings \cite{PromptOpt} or latents \cite{FreeDoM, DyMO, TFG, DOODL, UGD, DAS} through gradient descent. However, these test-time scaling methods suffer from inaccurate guidance and limited practicality, as single-image optimization on DiT-based models incurs a latency of several minutes or more.

\subsection{Hypernetworks}
Ha~\emph{et al.} \cite{Hypernetworks} proposed hypernetwork that predicts weights of a primary network, showing notable success in language modeling \cite{HyperInt, text2lora, DND}. For vision tasks, hypernetwork is applied across various domains, including segmentation \cite{Hyperseg}, image editing \cite{HyperStyle}, continue learning \cite{HyperCL}, 3D modeling \cite{Hyperpocket, Att3d}, personalization \cite{Hyperdreambooth, HyperNetField} and initial noise prediction for diffusion model \cite{HyperNoise, NoiseRefine}, among others.

\section{Problem Setup: Diffusion Model Alignment}
\subsection{Preliminary on Score-based Generative Models}
\label{sec:preliminary}

Diffusion models \cite{DDPM} capture a data distribution by learning to reverse a gradual noising process of applied to clean data. Given a data distribution $p_{\text{data}}(\bx)$, the {forward} process of a diffusion model \cite{DDPM, DDIM, SDE} progressively perturbs a clean sample $\bx_0\sim p_{\text{data}}(\bx)$ with Gaussian noise toward a Gaussian noise, following a stochastic differential equation (SDE) under certain conditions:
\begin{equation}
    \mathrm{d} \mathbf{x}_t = \mathbf{f}(\mathbf{x}_t)\,\mathrm{d}t + g_t\, \mathrm{d} \mathbf{w} ,
\end{equation}
where $t \in [0, T]$, $\mathbf{w}$ is a standard Wiener process, $\mathbf{f}(\mathbf{x}_t)$ and $g_t$ denote drift and diffusion coefficients, respectively \cite{SDE}. 

By running the above process backwards starting from $\bx_T\sim \mcN(0, \bI)$, we obtain a data generation process through the reverse SDE:
\begin{equation}
\label{eq:sde}
    \mathrm{d} \mathbf{x}_t = \bigl[\mathbf{f}(\mathbf{x}_t) - g^2_{t} \nabla_{\mathbf{x}_{t}} \log p_{t}(\mathbf{x}_{t})\bigr]\, \mathrm{d}t + g_t \, \mathrm{d} \mathbf{w} ,
\end{equation}
where $p_{t}(\mathbf{x}_{t})$ denotes the marginal distribution of $\mathbf{x}_t$ at time $t$. The score function $\nabla_{\mathbf{x}_{t}} \log p_{t}(\mathbf{x}_{t})$ can be estimated by training a model $\mathbf{s}_{\theta}(\mathbf{x}_t , t)$ \cite{SDE,TrainSDE}: 
\begin{equation}
\min_{\theta} \mathbb{E}_{t,\bx_0,\bx_t} \left\{ 
\lambda(t) ~ \| 
            \mathbf{s}_{{\theta}}(\mathbf{x}_t , t) - \nabla_{\mathbf{x}_t} \log p_t(\mathbf{x}_t | \bx_0 ) 
        \|_2^2 
\right\},
\end{equation}
where $\lambda(t)$ is the weight function, $\bx_0\sim p_\text{data}(\bx)$, $p_t(\bx_t | \bx_0)$ is a transition density in Gaussian, and $\bx_t\sim p_t(\bx_t | \bx_0)$. The approximated $\bs_{\theta}(\cdot)$ defines a learned distribution $p_\theta(\cdot)$. 

The score-based model unifies the formulations of diffusion models \cite{DDIM,DDPM} and flow matching models \cite{rectified_flow,SD3,FlowGRPO}, where the sample trajectories of $\bx_t$ are generated through a stochastic or ordinary differential equation (SDE or ODE) \cite{SDE}. 
For clarity and simplicity, we focus on diffusion models in the following presentation without loss of generality. Under this unified formulation, we can naturally generalize our analyses and approach to both diffusion \cite{DDIM} and flow-matching models \cite{rectified_flow}. More details can be found in Appendix.

\subsection{Aligning Diffusion Model with Reward}
\label{sec:conditional_score_diffusion}

\par{\textbf{Conditional diffusion models and score functions.}} 
We consider conditional diffusion models that learn a distribution $p_\theta(\bx|\bc)$ with $\bc$ denotes the conditioning variable. It is trained to generate samples through a reverse diffusion process via denoising a sampled noise $\bx_T$ under the control conditioning on $\bc$. For image generation, $\bc$ is the input prompts indicating user's instruction for the generated contents. 
We resort to discrete score-based model with the variance-preserving setting \cite{DDPM, DDIM} for better discussion, and its sampling formula is:
\begin{equation}
\label{eq:conditional_score_function}
  \bx_{t-1} = (1 + \frac{1}{2} \beta_t ) \bx_t + \beta_t \nabla_{\bx_t} \log p_{t}(\bx_t|\bc) + \sqrt{\beta_t} \, \boldsymbol{\epsilon},
\end{equation}
where $\bx_t \sim \mathcal{N}(\sqrt{\bar{\alpha}_t}\bx_0, \sigma_t^2 \bI)$, $\bar{\alpha}_t = \prod_{i=1}^{t}(1-\beta_i)$, $\sigma_t=\sqrt{1 - \bar{\alpha}_t}$, and $\beta_t$ is a linearly increasing noise scheduler. This iterative denoising process forms a \emph{trajectory} $\{\bx_t\}_{t=T}^0$ in the latent space, gradually transforming the noise $\bx_T$ into a clean sample $\bx_0$ reflecting the input prompt $\bc$.

\par{\textbf{Diffusion model alignment with reward.}} 
While existing T2I models demonstrate strong generative capabilities, their outputs often fall short of user expectations in visual quality and prompt semantic consistency. 
This limitation arises because the denoising objective optimizes for data reconstruction rather than human preference, and the learned score functions reflect the statistics of large-scale uncurated training data rather than the distribution of visually and semantically preferred outputs. Diffusion model alignment \cite{Alignment} is introduced to bridge this gap. 

Relying on human preference data \cite{HPS, hpsv3, PickAPic}, a reward model $R(\bx)$ can be obtained to capture human preference, \eg, aesthetic preference \cite{Aesthetic}. 
Conditioning on $\bc$, the reward model $R(\bx, \bc)$ can be formulated and assumed to partially capture the consistency between $\bx$ and $\bc$ as well as visual aesthetic preference. It can be learned explicitly from preference data or implicitly from raw data.

Given a learned $p_\theta(\bx|\bc)$ and a reward model, diffusion model alignment can be formulated as solving for a new tilted distribution \cite{DPO,DAS}:
\begin{equation}
\label{eq:RL_eq}
    p_{\theta,R} (\bx|\bc) = \frac{1}{\mcZ} p_\theta(\bx|\bc) \exp( \frac{R(\bx, \bc)}{\gamma}),
\end{equation}
where $\gamma$ is the KL regularization coefficient controlling the balance between reward maximization and consistency with the base model. Prevalent training-based alignment methods optimize the target rewards through RL \cite{SPO, DiffusionDPO, DiffusionKTO, DanceGRPO, MixGRPO} and direct backpropagation \cite{SRPO, AlignProp}. Although effective, these approaches risk reward over-optimization, leading to degraded generation diversity and reduced generalization across varied user inputs. Test-time scaling methods instead achieve alignment by applying guidance to steer intermediate latent states during inference. Since the generative distribution is characterized by the trajectory of $\bx_t$ in the sampling process, test-time alignment can be regarded as steering this trajectory to better match the desired conditional distribution $p_{\theta,R} (\bx|\bc)$. 

\begin{figure}[t]
  \centering
  \includegraphics[width=1.0\linewidth]{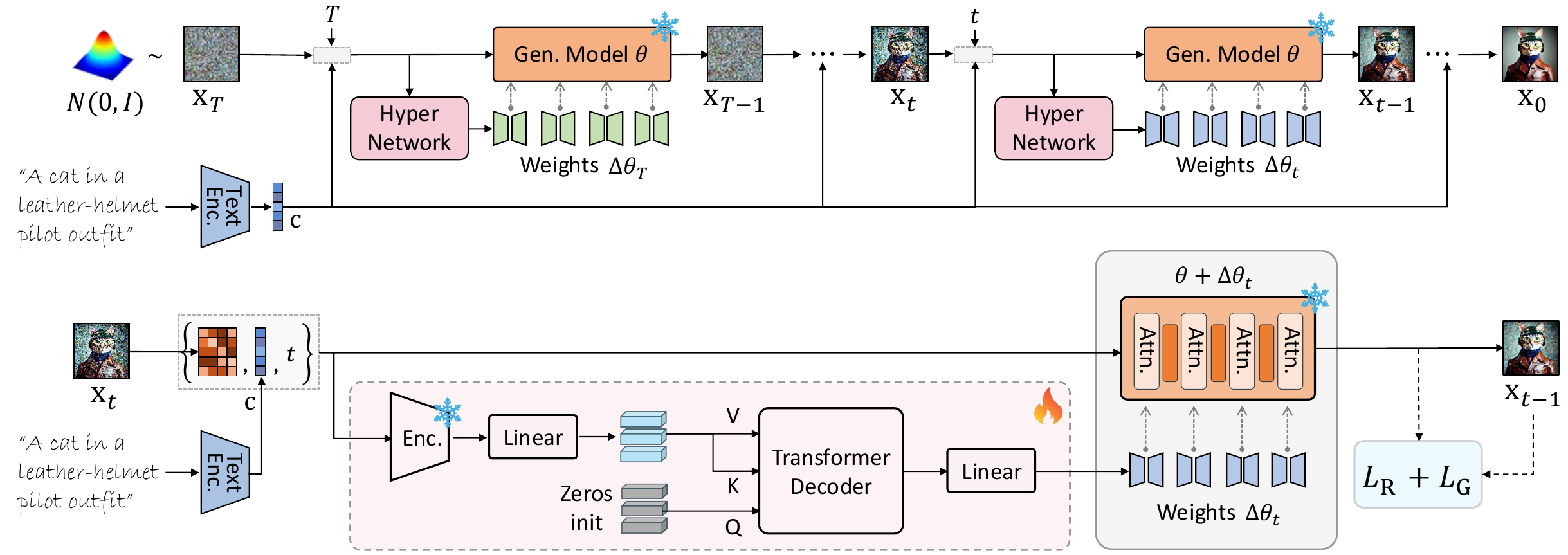}
  \caption{The framework of HyperAlign. Given a user prompt, the hypernetwork produces step-wise modulation weights $\Delta\theta_t$ that are injected into the generative model to steer the denoising trajectory (top). During training (bottom), the hypernetwork is optimized using the reward loss and the preference-regularization loss, enabling it to produce input-specific adjustments.}
\label{fig:framework} 
\end{figure}
\section{Alignment as Trajectory Steering and HyperAlign}
In this work, we propose HyperAlign, a learned hypernetwork for efficient and effective diffusion model alignment via test-time trajectory steering.

\subsection{Test-time Alignment with Diffusion Guidance}
\label{sec:modify_trajectory}
As discussed in Sec.~\ref{sec:conditional_score_diffusion}, test-time alignment methods adjust the generative trajectory conditioning on the input to better satisfy alignment objectives. Noise sampling methods \cite{BoN, EpsGreedy} search for favorable noise candidates based on reward feedback, but exploring the vast high-dimensional noise space is computationally expensive and prone to restricted reward optimization. 
Gradient-based diffusion guidance \cite{FreeDoM, DyMO,TFG} computes gradients from reward objectives to steer the trajectory via intermediate latent states, offering flexible control; however, it reacts only to instantaneous reward signals and incurs per-step gradient computational inference overhead.

\par{\textbf{Alignment via learned trajectory steering.}}
To achieve dynamic test-time alignment without intensive computation, we train a hypernetwork to generate prompt-specific and state-aware adaptations at each denoising step, steering the denoising trajectory toward target rewards. This amortizes costly test-time optimization into an efficient forward pass while enabling the model to learn state-aware alignment knowledge.

The idea of learning a hypernetwork for amortized test-time alignment of diffusion models has been explored in HyperNoise \cite{HyperNoise} by predicting input-conditioned initial noise. Training such a hypernetwork requires assigning all credit from supervision signals on the final generated images back to the initial noise prediction. It is restricted to distilled one-step models with post-hoc multi-step inference. In contrast, HyperAlign targets general multi-step diffusion and flow models. More discussions and analyses with HyperNoise are in Appendix. 

To learn amortized trajectory steering, we train a hypernetwork that absorbs reward-based diffusion guidance during training and operates as a simple feedforward process at inference. Before introducing HyperAlign, we first analyze how gradient-based test-time alignment methods adjust the generative trajectory. To achieve the reward-tilted distribution in Eq.~\eqref{eq:RL_eq}, gradient-based test-time alignment derives a score function with alignment scores as $\nabla_{\bx_t} \log \widetilde{p}_{\theta,R}(\bx_t|\bc) \approx \nabla_{\bx_t} \log p_{\theta}(\bx_t|\bc) + \lambda \cdot \nabla_{\bx_t} R(\bx_{0|t}, \bc)$, where $\nabla_{\bx_t} \log \widetilde{p}_{\theta,R}(\bx_t|\bc)$ denotes an approximated score of the aligned titled distribution, the first term on right corresponds to the conditional score from the given pre-trained model, requiring no extra optimization, and $\lambda$ denotes guidance strength. Focusing on the right-hand second term, it injects reward gradient into the denoising process:
\begin{equation}
\label{eq:final_guidance}
    \nabla_{\bx_t} R(\bx_{0|t}, \bc) = \frac{1}{\sqrt{\bar{\alpha}_t}} \cdot \frac{\partial R}{\partial \bx_{0\mid t}}\cdot \!\left(\mathbf{I}-\sqrt{1-\bar{\alpha}_t} \cdot \frac{\partial \boldsymbol{\epsilon}_{\theta}(\bx_t,t)}{\partial \bx_t} \right) ,
\end{equation}
where $\bx_{0|t}=\frac{1}{\sqrt{\bar{\alpha}_t}}(\bx_t-\sqrt{1-\bar{\alpha}_t}\boldsymbol{\epsilon}_{\theta}(\bx_t,t))$ is the estimated clean data \cite{chung2022diffusion} and the reward function is actually applied on the decoded image domain through the decoder. We omit the decoder notation for simplifying presentation. Substituting Eq.~\eqref{eq:final_guidance} into Eq.~\eqref{eq:conditional_score_function}, guidance-based methods achieve alignment by injecting reward-aware dynamics into each transition from $\bx_t$ to $\bx_{t-1}$, effectively steering the entire denoising trajectory. The adjustment in Eq.~\eqref{eq:final_guidance} is conditioned on the input $\bc$ (via $R(\cdot, \bc)$), the intermediate state (via $\bx_t$), and the initial noise (via $\bx_t,t=T$).

\begin{figure}[t]
  \centering
  \includegraphics[width=1.0\linewidth]{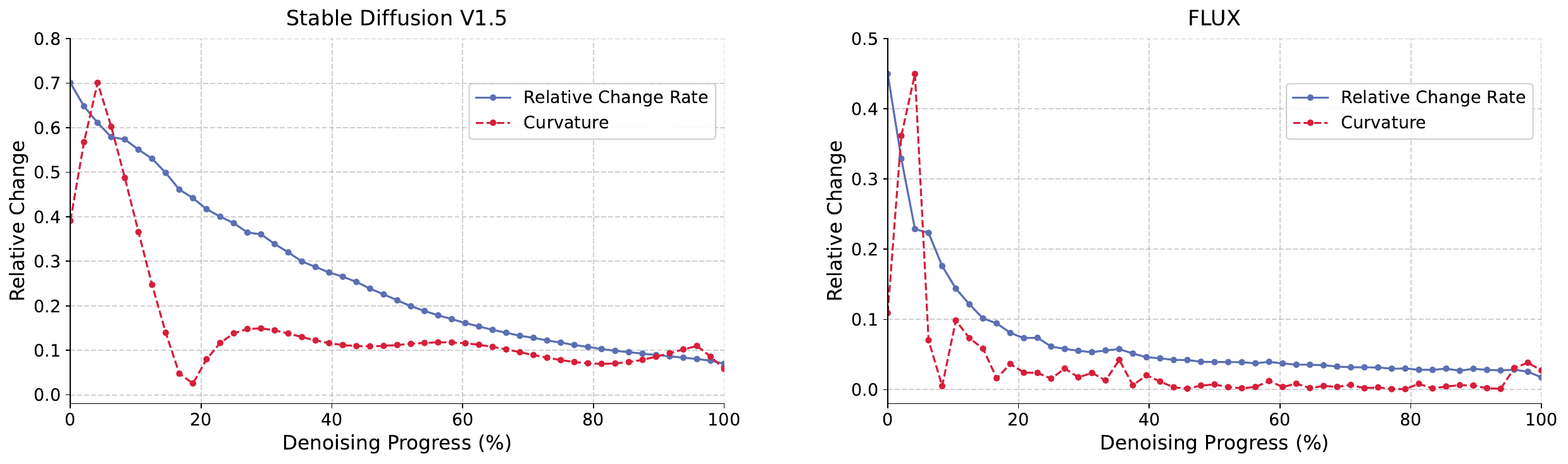}
  \caption{The prompt-invariant temporal dynamics of predicted clean data. Average over 1000 prompts. The relative change rate (blue) of both SD V1.5 and FLUX decreases monotonically, indicating a transition from coarse structural construction in the early stage to fine-grained detail refinement in later steps. The curvature (red) further exhibits distinct stage-wise patterns, reflecting different dynamic behaviors throughout the denoising process.}
\label{fig:curvature} 
\end{figure}

\subsection{HyperNetwork for Test-time Alignment}
To amortize test-time computation into feedforward operations, we learn a hypernetwork that produces trajectory adjustments conditioned on input prompts, initial noise, and intermediate latents, mirroring the functionality of Eq.~\eqref{eq:final_guidance} as discussed in Sec.~\ref{sec:modify_trajectory}. We train the hypernetwork with a reward objective to absorb reward signals into its learned parameters. Unlike fine-tuning-based alignment, which accommodates all intentions into the fixed set of parameters, HyperAlign is prompt-specific and state-aware, dynamically generating adaptive modulation parameters at each denoising step to steer the generation trajectory. 

\par{\textbf{Hypernetwork as a dynamic LoRA predictor.}}
We aim to learn a hypernetwork that takes $\bx_t$, $t$ and $\bc$ as input and outputs adjustments for each step of the generative process. 
A straightforward approach would be to learn a hypernetwork that directly generates an alignment score as a substitute for Eq.~\eqref{eq:final_guidance}, analogous to the original generative score $\nabla_{\bx_t} \log p_{\theta}(\bx_t|\bc)$. However, this requires the hypernetwork to match the full complexity of the score network, leading to prohibitive implementation costs. Tracing back to the idea of trajectory steering, we can treat the aligned score $\nabla_{\bx_t} \log \widetilde{p}_{\theta,R}(\bx_t|\bc)$ as a variation of the original pre-trained model's score. We thus design the hypernetwork to generate adjustments to the original diffusion model that produces the score. Specifically, the hypernetwork generates adjustments to the network parameters $\theta$ of the original generative model by producing lightweight low-rank adapters (LoRA) for $\theta$. 

We divide hypernetwork architecture for generating LoRA weights into two main components: \emph{perception encoder} and \emph{transformer decoder}, as shown in Fig.~\ref{fig:framework}. Concretely, the inputs temporal latent $\mathbf{x}_t$, timestep $t$ and prompt embeddings $\bc$ are first passed into perception encoder, which consists of downsampling blocks from the pretrained U-Net of the generative model \cite{SD}. The pretrained U-Net carries rich priors, making it a natural encoder to capture semantic representations across diverse input combinations. The encoded features are then projected as context feature through a linear layer. For weights generation, we adapt a transformer decoder \cite{Hyperdreambooth} (consisting of standard transformer blocks) with the inputs of zero-initialized tokens and encoded context features, followed by a linear layer maps the decoded features into LoRA weights: 
\begin{equation}
    \Delta {{\theta}_t} = h_{\psi}(\mathbf{x}_t, \bc, t) ,
    \label{eq:hyper}
\end{equation}
where $\psi$ denotes the parameters of hypernetwork $h_{\psi}$. Detailed hypernetwork architecture are provided in Appendix. Temporally, integrating the generated LoRA weights into the original model parameters yields a input-and-step-specific score function $\mathbf{s}_{\theta+\Delta\theta_t}$ (with an abuse of notation $+$), thereby modifying the underlying denoising trajectory. The generated LoRA weights are specified for inputs and adaptive for the dynamics in trajectories. More analyses are in Appendix. 

\par{\textbf{HyperAlign with efficiency augmentation.}}
By default, the hypernetwork design in Eq.~\eqref{eq:hyper} is applied at all generation steps starting from initial step $T$ (termed as HyperAlign-S). We develop two more variants for balancing and augmenting inference efficiency. (1) HyperAlign-I generates LoRA weights once at the beginning of the denoising process and uses them for all subsequent denoising steps. (2) A piece-wise variants, HyperAlign-P, produces new LoRA weights at a few key timesteps. Denoising behaviors vary across different stages but are similar within each stage \cite{FreeDoM, DyMO}, as reflected by the relative $\ell_1$ distance between adjacent predicted clean latents in Fig.~\ref{fig:curvature}. Early steps exhibit larger latent changes (coarse structure construction), while later steps become more stable (fine-grained appearance refinement). The observations inspire us that the denoising steps with similar behaviors can be grouped into a single segment and share identical LoRA weights. To determine stage boundaries, we compute the curvature rate and identify $M$ stage transition points where the curvature trend changes (\eg, 0\%, 5\%, 20\% for SD V1.5 and 0\%, 5\%, 10\% for FLUX). The hypernetwork regenerate LoRA weights only at these transition points, enabling adaptive modulation of the diffusion process with significantly fewer updates than HyperAlign-S, resulting in efficiency–performance trade-off.

\subsection{Training HyperAlign}
The hypernetwork is learned in a post-training process with reward scores as supervision, optimizing for generating LoRA for diffusion trajectory alignment while keeping the pre-trained model frozen. By maximizing the reward signals, the model is encouraged to generate intermediate predictions with higher conditional likelihood, thereby aligning the latent trajectory with the true conditional distribution:
\begin{equation}
\begin{aligned}
    \mathcal{L}_{\text{R}} 
    &= - \mathbb{E}_{\bc \sim \mathcal{D}, \bx_{0|t} \sim p_{\theta+\Delta\theta_t}(\bx_0|\bx_t, c)} \!\left[ R(\bx_{0|t}, \bc) \right],
\end{aligned}
\end{equation}
where $\bx_t$ is sampled from the LoRA-modulated denoising trajectories with the random initial noise $\bx_T \sim \mathcal{N}(\mathbf{0}, \bI)$. The hypernetwork is trained end-to-end by backpropagating supervision signals directly through the frozen pre-trained model and dynamically on-the-fly generated LoRA weights, requiring no pre-trained LoRA targets as supervision. 

\par{\textbf{Regularization on reward optimization.}}
While maximizing reward objective drives the model to produce high-reward, condition-aligned latent states, it also exposes two key challenges: (1) inaccurate reward signals due to the blurriness of predicted clean data in early denoising stages, and (2) the risk of over-optimization, where aggressive reward maximization leads to reward hacking or degraded visual fidelity. To further mitigate this issue, we incorporate a regularization loss on alignment process, while preserving generation quality:
\begin{equation}
\hspace{-0.3cm}
\begin{aligned}
\mathcal{L}_{\text{G}} = 
\mathbb{E}_{(\bx_0, \bc) \sim \mathcal{D}, \bx_t \sim p_t(\bx_t|\bx_0)} 
\Big[
    \eta_t
    \big\|
        \nabla_{\bx_t}\log p_{\theta+\Delta\theta_t}(\bx_t|\bc)
        - \nabla_{\bx_t}\log p_{t}(\bx_t|\bx_0)
    \big\|_2^2
\Big],
\end{aligned}
\end{equation}
where $\eta_t$ denotes the hyperparameter, preferred data $\mathbf{x}_0$ sampled from preference dataset $\mathcal{D}$ and $\mathbf{x}_t$ is sampled from forward diffusion process $p_t(\mathbf{x}_t|\mathbf{x}_0)$. We encourages the predicted denoising conditional score to remain close to true marginal score of preferred data, prevent reward hacking by suppressing noisy rewards from early-step blur and preserving visual fidelity in later denoising stages.
The final learning objective for the hypernetwork optimization can be described as follows:
\begin{equation}
\psi^{*} = \arg\min_{\psi} \left\{\mathcal{L}_{\text{R}} + \mathcal{L}_{\text{G}} \right\}.
\end{equation}

As mentioned, HyperAlign is not limited to diffusion models and is also compatible with flow-matching models (\eg, FLUX in experiments). More details are in Appendix.

\begin{figure}[t]
  \centering
  \includegraphics[width=1.0\linewidth]{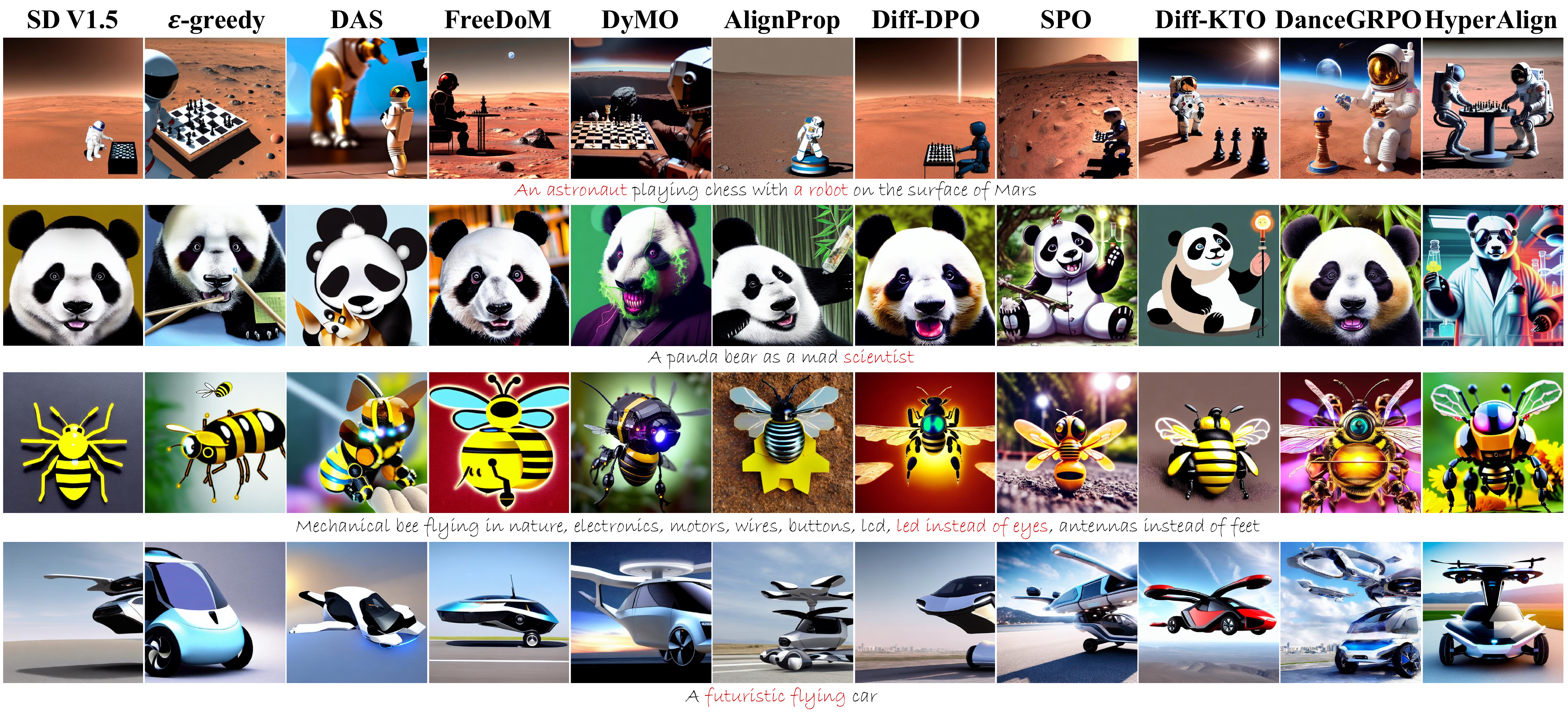}
  \caption{Qualitative comparison based on SD V1.5 backbones.}
\label{fig:sd_compare} 
\end{figure}

\section{Experiments}
\label{sec:experiments}
In this section, we conduct comprehensive experimental evaluations to verify the effectiveness and efficiency of our method. We flexibly apply it across various generative paradigms and validate its performance through comparisons with existing state-of-the-art approaches. Moreover, ablation studies are performed to substantiate the contribution of each component in our designs.

\subsection{Experimental Setting}
\par
\textbf{Implementation Details.} 
We employ SD V1.5 \cite{SD} and FLUX.1.Dev \cite{FLUX} as base models, paired HPSv2 \cite{HPS} as the reward model. The preferred data used for regularization loss originates from Pick-a-Pic \cite{PickAPic} and HPD \cite{HPS}. All our experiments uses four NVIDIA H100 GPUs. We provide the detailed hypernetwork configurations, training setups and cost in Appendix.

\par
\textbf{Datasets and Metrics.} 
We evaluate our method on four datasets: 1K prompts from Pick-a-Pic \cite{PickAPic}, 2K from GenEval \cite{geneval}, 500 from HPD \cite{HPS}, and 1K from Partiprompt \cite{Partiprompt}. We choose six AI feedback models to assess the generated image quality: PickScore \cite{PickAPic} and ImageReward (IR) \cite{ImageReward} for general human preference, HPSv2 \cite{HPS}, CLIP \cite{clip} and GenEval Scorer \cite{geneval} for prompt alignment, Aesthetic Predictor \cite{Aesthetic} for visual appeal. All test images are produced with 50 denoising steps across different methods for a fair comparison, where the CFG scale coefficient is set to 7.5 for SD V1.5-based backbones and 3.5 for FLUX-based backbones. {For all metrics, higher values indicate better performance.}

\begin{table*}[t]
\centering
\begin{minipage}[t]{0.48\columnwidth}
\centering
\renewcommand{\arraystretch}{1.0}
\caption{Comparison of AI feedback on SD V1.5-based methods.}
\resizebox{1.0\linewidth}{!}{
\setlength{\tabcolsep}{1.5mm}{
\begin{tabular}{c|ccccc|c}
\hline
Method & Aes & Pick & IR & CLIP & HPS & Time  \\ 
\hline
SD V1.5        & 5.443 & 20.66 & 0.1111 & 0.2745 & 0.2500 & 3s \\
\hline
$\varepsilon$-greedy        & 5.509 & 21.02 & 0.4906 & 0.2821 & 0.2737 & 250s \\
DAS   & 5.231 & 19.43 & -0.4524 & 0.2420 & 0.2387  & 25s\\
FreeDoM   & 5.690 & 21.16 & 0.4419 & 0.2822 & 0.2779 & 120s \\
DyMO   & 5.723 & 21.89 & 0.6944 & \textbf{0.2901} & \underline{0.2944} & 162s \\
\hline
AlignProp     & 5.409 & 20.58 & 0.1345 & 0.2750 & 0.2501 & 3s \\
Diffusion-DPO         & 5.519 & 21.01 & 0.3201 & 0.2795 & 0.2601 & 3s \\
SPO        & 5.664 & 21.15 & 0.2978 & 0.2599 & 0.2741 & 3s \\
Diffusion-KTO         & 5.660 & 21.16 & 0.6970 & 0.2809 & 0.2824 & 3s \\
DanceGRPO   & 5.711 & 21.15 & 0.6886 & 0.2748 & 0.2925 & 3s \\ \hline
\rowcolor{tabcolor} HyperAlign-I  & 5.791 & \underline{21.94} & 0.5947 & \underline{0.2870} & 0.2877  & 3s \\
\rowcolor{tabcolor} HyperAlign-P  & \textbf{5.878} & 21.89 & \underline{0.6979} & 0.2823 & 0.2885  & 4s\\
\rowcolor{tabcolor} HyperAlign-S  & \underline{5.824} & \textbf{22.01} & \textbf{0.7731} & 0.2851 & \textbf{0.2957} & 5s \\
\hline
\end{tabular}}}
\label{tab:sd_compare_pick}
\end{minipage}
\hfill
\begin{minipage}[t]{0.48\columnwidth}
\centering
\renewcommand{\arraystretch}{1.13}
\caption{ Comparison of AI feedback on FLUX-based methods.}
\resizebox{1.0\linewidth}{!}{
\setlength{\tabcolsep}{1.5mm}{
\begin{tabular}{c|ccccc|c}
\hline
Method & Aes & Pick & IR & CLIP & HPS & Time \\ 
\hline
FLUX.1.Dev        & 6.182 & 22.26 & 1.023 & 0.2651 & 0.3072 & 15s \\
\hline
BoN         & 6.016 & 22.26 & 1.124 & 0.2668 & 0.3114 & 300s \\
$\varepsilon$-greedy & 6.166 & \underline{22.30} & 1.077 & 0.2662 & 0.3115 & 1100s \\
FreeDoM  & 5.933 & 21.67 & 0.842 & \underline{0.2672} & 0.2970 & 240s \\
DyMO  & 6.165 & 21.74 & 1.062 & \textbf{0.2714} & 0.3004 & 300s \\
\hline
DanceGRPO   & 6.706 & 22.08 & 1.135 & 0.2341 & 0.3527 & 15s \\
MixGRPO     & 6.760 & 22.25 & 1.240 & 0.2499 & 0.3460 & 15s \\
SRPO        & 6.061 & 22.16 & 0.833 & 0.2624 & 0.2917 & 15s \\ \hline
\rowcolor{tabcolor} HyperAlign-I  & 6.733 & 22.18 & 1.242 & 0.2511 & 0.3529 & 16s \\
\rowcolor{tabcolor} HyperAlign-P  & \underline{6.769} & 22.13 & \textbf{1.280} &  0.2506 & \underline{0.3530} & 17s \\
\rowcolor{tabcolor} HyperAlign-S  & \textbf{6.853} & \textbf{22.37} & \underline{1.251} & 0.2602 & \textbf{0.3611} & 20s \\
\hline
\end{tabular}}}
\label{tab:flux_compare_pick}
\end{minipage}
\end{table*}

\begin{figure}[t]
  \centering
  \includegraphics[width=1.0\linewidth]{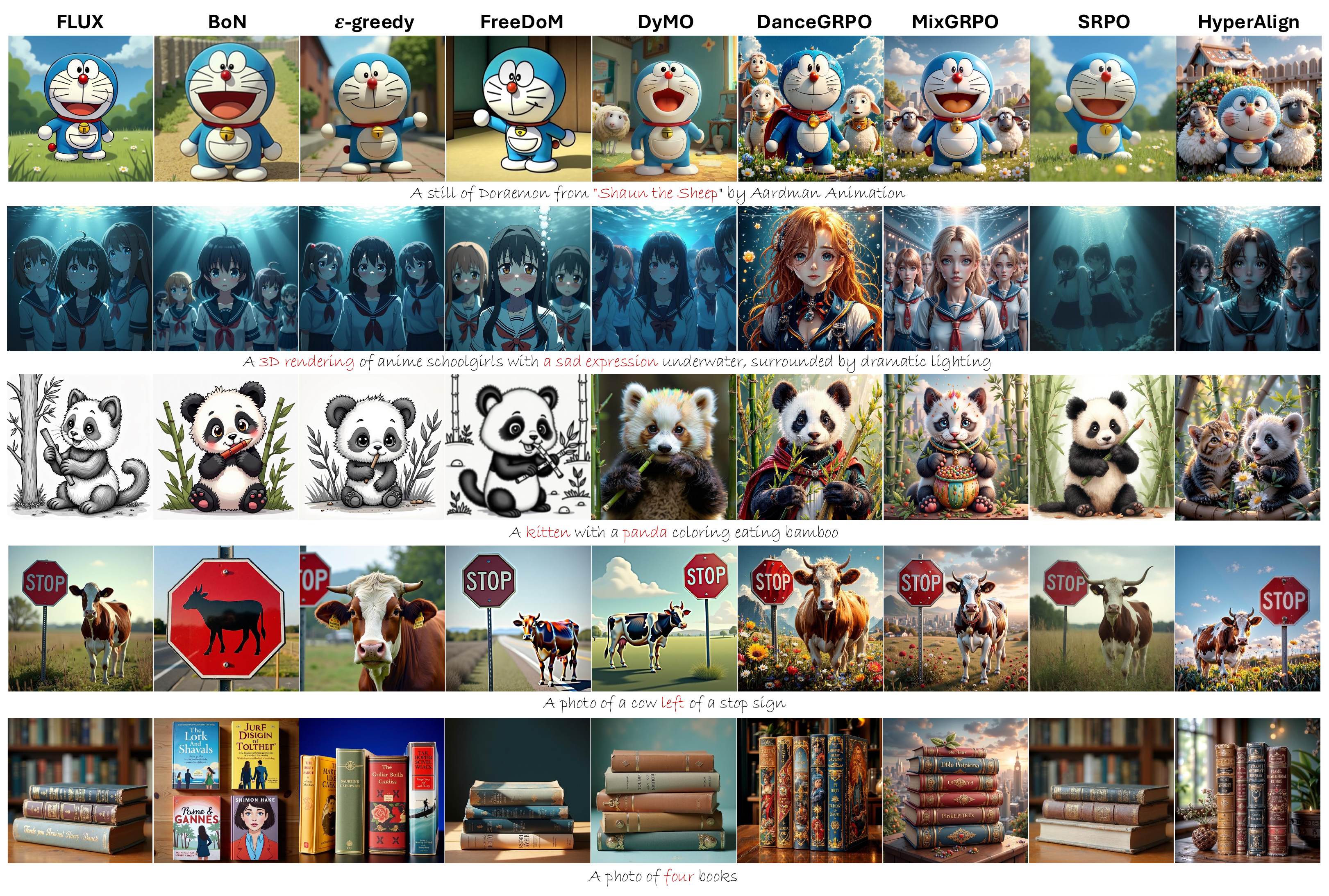}
  \caption{Qualitative comparison examples based on FLUX backbones.}
\label{fig:flux_compare} 
\end{figure}

\subsection{Comparison with Existing Methods}
\label{sec:Comparison_w_Existing}

For comprehensive assessment, we compare our method with training-based and test-time alignment methods. The former covers RL methods with direct reward backpropagation \cite{AlignProp, SRPO} and policy optimization paradigms \cite{DiffusionDPO, SPO, DiffusionKTO, DanceGRPO}. The latter comprises reward gradient-based guidance methods \cite{FreeDoM, DyMO, DAS} and noise candidate search strategies \cite{BoN, EpsGreedy}. For noise sampling methods, we follow the original configuration by setting the number of noise candidates to 20 for BoN \cite{BoN}, and using 20 local search iterations with 4 noise candidates for $\varepsilon$-greedy \cite{EpsGreedy}.

\subsubsection{Quantitative Analysis}
To objectively evaluate the performance of our method, we conduct a quantitative comparison on the Pick-a-Pic dataset. For fair comparison, all alignment methods only use the HPSv2 scorer \cite{HPS} as the reward model. The results are organized in Tab.~\ref{tab:sd_compare_pick} for SD V1.5-based backbones and Tab.~\ref{tab:flux_compare_pick} for FLUX-based backbones, respectively. It is observed that our method effectively achieve alignment and outperform the previous methods by adjusting the generation trajectory step by step. The other two variants of our method also keep competitive performance with faster inference. More quantitative results across various benchmarks are provided in Appendix.

\subsubsection{Qualitative Evaluation}
We further provide some visual results of the generated images in Fig.~\ref{fig:sd_compare} for SD V1.5-based backbones and Fig.~\ref{fig:flux_compare} for FLUX-based backbones. Qualitative comparison demonstrates the superior visual quality of our method, consistently producing images with coherent layouts, semantically rich content aligned with the prompts, and aesthetics. More visualization results can be found in Appendix.

\begin{figure}[t]
\centering
 \includegraphics[width=1.0\linewidth]{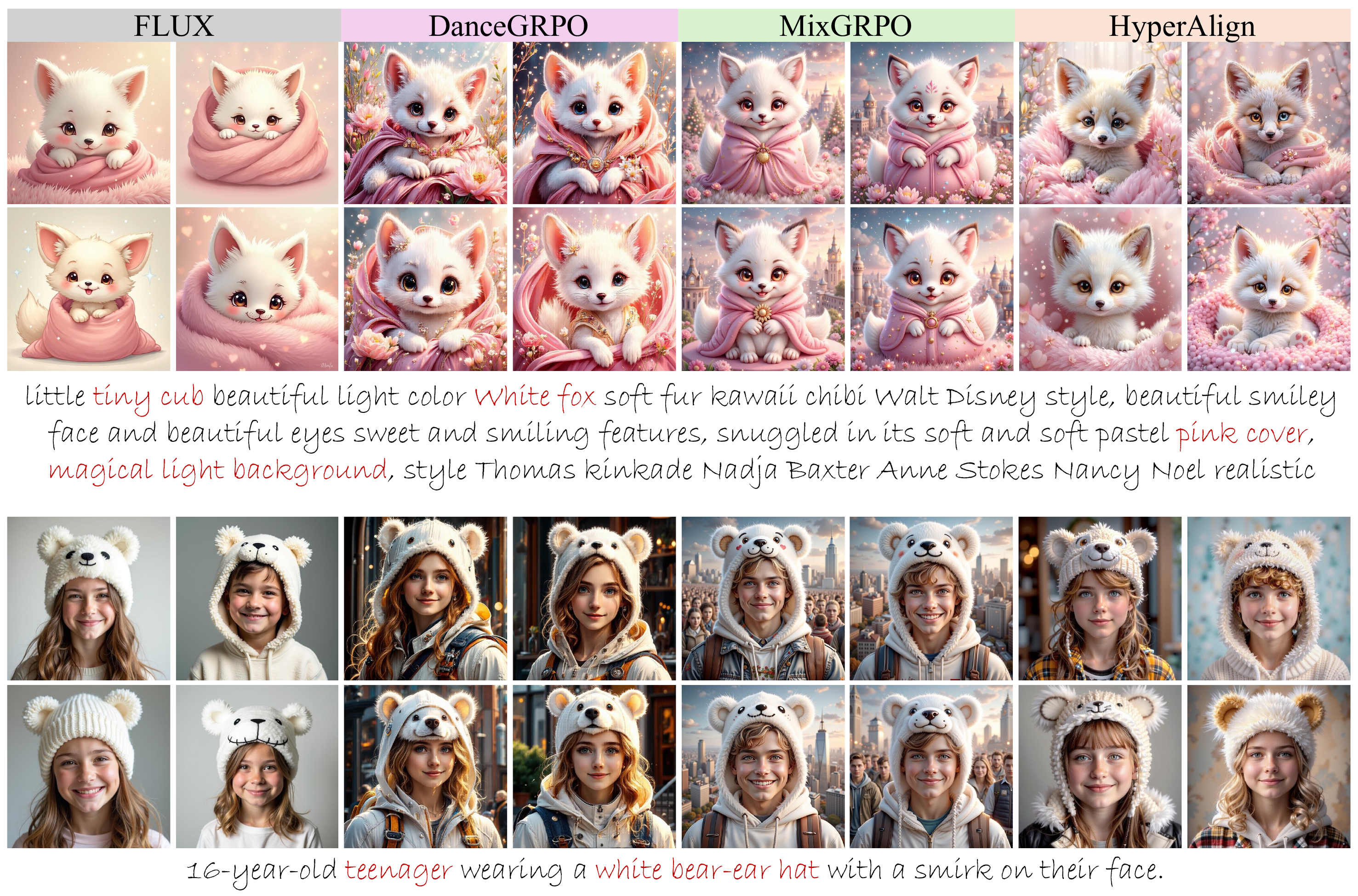}
\caption{Diversity comparison based on FLUX backbones.}
\label{fig:diversity}
\end{figure}

\subsubsection{Diversity.} 
\begin{wraptable}{R}{0.36\textwidth}
  \centering
  \renewcommand{\arraystretch}{1.0}
  \caption{Average similarity scores.}
  \label{tab:Avg_Similarity}
  \resizebox{0.35\textwidth}{!}{
  \setlength{\tabcolsep}{3.0mm}{
    \begin{tabular}{c|cc}
      \hline
      Method & LPIPS $\uparrow$ & DINO $\downarrow$ \\
      \hline
      FLUX.1.Dev & 0.538 & 0.746 \\
      DanceGRPO  & 0.466 & 0.774 \\
      MixGRPO    & 0.390 & 0.839 \\
      HyperAlign & 0.497 & 0.751 \\
      \hline
  \end{tabular}}}
  \vspace{-0.1cm}
\end{wraptable}
We also investigate the impact of the diversity of the generated outputs across different methods. Specifically, for each prompt in the HPD benchmark, we generate 50 images using different random seeds. The average similarity of different images for the same prompt are measured using similarities from LPIPS \cite{lpips} and DINOv2 \cite{dinov2} embeddings. As shown in Tab.~\ref{tab:Avg_Similarity}, HyperAlign preserves the diversity of generated outputs. We also provide the visual results in Tab.~\ref{tab:flux_compare_pick}, where the outputs of previous methods \cite{DanceGRPO, MixGRPO} tend to collapse toward a single style or even a single identity.

\subsubsection{Efficiency} 
We report the average inference time per image in Tab.~\ref{tab:sd_compare_pick} and Tab.~\ref{tab:flux_compare_pick} for SD V1.5- and FLUX-based backbones, respectively.
By amortizing iterative test-time computations into feedforward hypernetwork passes via training, HyperAlign achieves input-state-adaptive alignment efficiently. Training the hypernetwork takes 8 and 12 GPU hours for SD V1.5 and FLUX, respectively. For SD V1.5, it is roughly equivalent to generating 178 images with DyMO \cite{DyMO} or 240 with FreeDoM \cite{FreeDoM} - it is a modest investment for scalable deployment. For applications generating images at scale, this yields substantial long-term savings, same as all training-based methods. Compared to other training-based methods, HyperAlign obtains better performance with only minimal overhead from LoRA weight generation and loading. Further details are provided in the Appendix. 

\subsubsection{User Study}
We conduct a subjective user study on FLUX-based backbones by randomly sampling 100 unique prompts from the HPD benchmark \cite{HPS} and generating the corresponding images using our method and several state-of-the-art baselines \cite{DyMO, SRPO, DanceGRPO, MixGRPO}. A total of 100 participants are invited to evaluate each comparison group by selecting the most favorable image across three criteria: Q1 General Preference (Which image do you prefer given the prompt?), Q2 Visual Appeal (Which image is more visually appealing?), Q3 Prompt Alignment (Which image better fits the text description?), shown in Appendix. Fig.~\ref{fig:user_study} shows approval percentage of each method in three aspects, demonstrating our method outperforms the previous alignment methods on human feedback.

\subsection{Ablation Study}
\begin{figure}[!t]
\centering
\begin{minipage}[t]{0.48\columnwidth}
\vspace{0cm}
\centering
\includegraphics[width=1.0\linewidth]{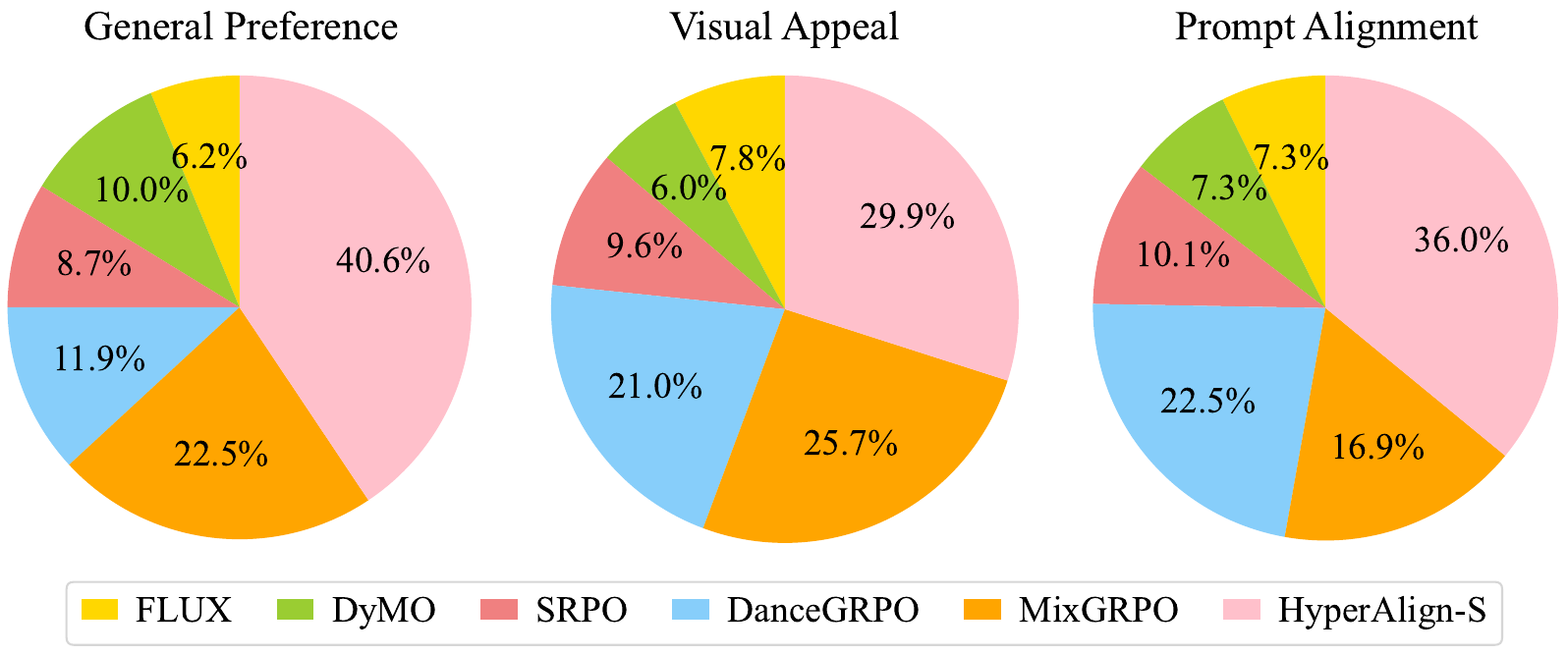}
\captionof{figure}{User study results.}
\label{fig:user_study} 
\end{minipage}
\hfill
\begin{minipage}[t]{0.48\columnwidth}
\vspace{-0cm}
\centering
\captionof{table}{Ablation study results.}
\vspace{0cm}
\renewcommand{\arraystretch}{1.0}
\resizebox{1.0\linewidth}{!}{
\setlength{\tabcolsep}{1.5mm}{
\begin{tabular}{c|ccccc}
\hline
Method & Aes & Pick & IR & CLIP & HPS  \\ 
\hline
HPSv2+Pick-a-Pic & 5.824 & 22.01 & 0.7731 & 0.2851 & 0.2957 \\
HPSv2+HPD & 5.852 & 21.24 & 0.6627 & 0.2765 & 0.2833 \\
\hline
PickScore+HPD & 5.871 & 21.45 & 0.6317 & 0.2734 & 0.2837 \\
PickScore+Pick-a-Pic & 5.720 & 21.73 & 0.6832 & 0.2735 & 0.2842 \\
\hline
Only Pick-a-Pic & 5.607 & 20.88 & 0.5615 & 0.2745 & 0.2743 \\
Only HPD & 5.413 & 20.37 & 0.3314 & 0.2734 & 0.2551 \\
Only HPSv2  & 5.702 & 20.43 & 0.7629 & 0.2496 & 0.3128 \\
Only PickScore  & 6.000 & 21.45 & 0.6450 & 0.2498 & 0.2796 \\
\hline
\end{tabular}}}
\label{tab:ablation_study}
\end{minipage}
\end{figure}

To better understand the contributions of each component in our framework, we conduct a series of ablation studies on SD V1.5 \cite{SD} under the HyperAlign-S configuration unless otherwise specified. The results are summarized in Tab.~\ref{tab:ablation_study}.

\par
\textbf{Effect of preference data for regularization loss $\mathcal{L}_{\text{G}}$.} Our default configuration adopts HPSv2 as the reward model and Pick-a-Pic as the preference dataset for regularization. When replacing Pick-a-Pic with HPD while keeping HPSv2 fixed, our method still achieves strong performance, demonstrating the robustness and effectiveness of our method.

\par
\textbf{Effect of reward–regularization configurations.} Beyond HPSv2, we combine PickScore and different preference datasets to optimize the hypernetwork. All combinations lead to consistently solid outcomes, verifying that HyperAlign can adapt to different reward and regularization sources. Our default choice, HPSv2 leans toward text–image alignment while Pick-a-Pic dataset favors visual appeal, provides balanced supervision that yields stronger overall improvements across metrics.

\par
\textbf{Effect of reward loss $\mathcal{L}_{\text{R}}$.} We further examine the influence of the reward loss by supervised fine-tuning using only preference data (Pick-a-Pic and HPD) and optimization using only reward signals (HPSv2 and PickScore). Results show that supervised fine-tuning with preference data alone yields marginal gains. Reward-only optimization boosts most preference scores but severely degrades CLIP, indicating clear reward over-optimization.

\section{Conclusion}
\label{sec:conclusion}
We propose HyperAlign, a hypernetwork-based framework for efficient and effective test-time alignment of generative models. HyperAlign dynamically generates low-rank modulation weights across denoising steps, enabling trajectory-level alignment guided by reward signals. Its variants provide flexible trade-offs between computational efficiency and alignment precision. Extensive experiments on both diffusion and rectified flow backbones show that HyperAlign delivers superior semantic consistency and aesthetic quality compared to existing fine-tuning and test-time alignment approaches. 
In the future, we aim to further enhance dynamic adaptation while developing more lightweight hypernetwork designs to improve efficiency and scalability.

\bibliographystyle{splncs04}
\bibliography{main}

\clearpage
\newpage
\beginappendix

\section{More Details of HyperAlign with Flow-Matching Models}

HyperAlign is designed to be model-agnostic, applicable to both diffusion and flow-matching models. The main paper formalizes the method using diffusion models for simplifying presentation. The core mechanism of steering alignment trajectory towards target reward can be naturally transferred to flow-matching models \cite{FLUX}, as demonstrated in the experiments with FLUX. While the connections between diffusion and flow-matching models have been established and unified formulations have been presented in prior work \cite{DanceGRPO, Adjoint_matching}, we provide below derivations of trajectory alignment with reward tailored to flow-matching models, to ensure clarity and reproducibility.

\par
\textbf{Conditional flow-matching models and score functions.}
Unlike reverse SDE in Eq.~\eqref{eq:sde}, a deterministic reverse probability flow ODE \cite{SDE} takes the following form without randomness:
\begin{equation}
\label{eq:ode}
    \mathrm{d} \mathbf{x}_t = \bigl[\mathbf{f}(\mathbf{x}_t) - \frac{1}{2}g^2_{t} \nabla_{\mathbf{x}_{t}} \log p_{t}(\mathbf{x}_{t})\bigr]\, \mathrm{d}t .
\end{equation}

Flow-matching models \cite{rectified_flow, FLUX, SD3} adopt this deterministic transport perspective, but parameterize the entire drift as a velocity field $v_t(\bx_t)$:
\begin{equation}
    \mathrm{d} \mathbf{x}_t = v_t(\bx_t) \mathrm{d}t ,
\end{equation}
where $v_t(\bx_t)$ implicitly contains the score $\nabla_{\mathbf{x}_{t}} \log p_{t}(\mathbf{x}_{t})$. To explore the incorporation of reward guidance into flow-matching models, we need to recover the score from $v_t$. In this section, to formulate the flow matching process, we let $\bx_0$ denote a data sampled from the true distribution and $\bx_1$ be a noise sample \cite{MixGRPO,FlowGRPO}. Specifically, let $\bx_0 \sim p_\text{data}(\bx)$ and $\bx_1 \sim \mathcal{N}(0, \mathbf{I})$. Then the forward process can be formulated as a linear interpolation:
\begin{equation}
    \bx_t = \alpha_t \bx_0 + \beta_t \bx_1 ,
\end{equation}
where $\alpha_t=1-t$, $\beta_t = t$ and $t \in [0,1]$. Under this construction, we have the distribution $\bx_t \sim \mathcal{N}(\alpha_t \bx_0, \beta_t^2 \mathbf{I})$, yielding the marginal score:
\begin{equation}
\begin{aligned}
    \nabla_{\mathbf{x}_{t}} \log p_{t}(\bx_t) = \mathbb{E} \!\left[ \nabla_{\mathbf{x}_{t}} \log p_{t|0} (\bx_t|\bx_0) | \bx_t \right] = - \frac{1}{\beta_t} \mathbb{E} \!\left[ \bx_1|\bx_t  \right] .
\end{aligned}
\end{equation}

For the velocity field $v_t(\bx_t)$, we derive:
\begin{equation}
\begin{aligned}
\label{eq:vt_score}
    v_t(\bx_t) 
    &= \mathbb{E}\!\left[\, \dot{\alpha}_t \bx_0 + \dot{\beta}_t \bx_1 \, | \, \bx_t \right] \\
    &= \dot{\alpha}_t \mathbb{E}[\bx_0 | \bx_t] + \dot{\beta}_t \mathbb{E}[\bx_1 | \bx_t ] \\
    &= \dot{\alpha}_t \mathbb{E}\!\left[\, \frac{\bx_t - \beta_t \bx_1}{\alpha_t} \,|\, \bx_t \right]
       + \dot{\beta}_t \mathbb{E}[\bx_1 | \bx_t] \\
    &= \frac{\dot{\alpha}_t}{\alpha_t} \bx_t 
       - \frac{\dot{\alpha}_t \beta_t}{\alpha_t} \mathbb{E}[\bx_1 | \bx_t]
       + \dot{\beta}_t \mathbb{E}[\bx_1 | \bx_t ] \\
    &= \frac{\dot{\alpha}_t}{\alpha_t} \bx_t 
       - \left( \beta_t \dot{\beta}_t 
       - \frac{\dot{\alpha}_t \beta_t^{2}}{\alpha_t} \right)
       \nabla_{\mathbf{x}_{t}} \log p_{t}(\bx_t) .
\end{aligned}
\end{equation} 

To extend this to conditional generation, we replace the unconditional score $\nabla_{\mathbf{x}_{t}} \log p_{t}(\bx_t)$  with the conditional score  $\nabla_{\mathbf{x}_{t}} \log p_{t}(\bx_t|\bc)$, where $\bc$ denotes the conditioning variable (\eg, text prompts). Applying discretization to $v_t = \frac{\mathrm{d} \bx_t}{\mathrm{d} t}$ yields the conditional latent update rule:
\begin{equation}
\label{eq:ode_score_guidance}
    \bx_{t+\Delta t} = (1-\frac{\Delta t}{1-t})\bx_t - \frac{t \, \Delta t}{1-t} \nabla_{\mathbf{x}_{t}} \log p_{t}(\bx_t | \bc).
\end{equation}

Similar to Eq.~\eqref{eq:conditional_score_function}, this iterative sampling process also forms a trajectory $\{\bx_t\}_{t=1}^0$ in the latent space, gradually transforming the noise $\mathbf{x}_1$ into a clean sample $\mathbf{x}_0$.  

\par
\textbf{Test-time alignment with reward-based guidance.}
Given a learned $p_{\theta}(\bx_t|\bc)$ and a reward model $R$, test-time alignment aims to steer the produced trajectory from $\bx_1$ to $\bx_0$, targeting the generated samples conform to the reward-titled distribution $p_{\theta,R} (\bx|\bc)$, as discussed in Sec.~\ref{sec:conditional_score_diffusion} and Sec.~\ref{sec:modify_trajectory}. To achieve this in flow-matching models \cite{FLUX, SD3}, we still focus on reward-guidance term of the alignment score $\nabla_{\bx_t} \log \widetilde{p}_{\theta,R}(\bx_t|\bc) \approx \nabla_{\bx_t} \log p_{\theta}(\bx_t|\bc) + \lambda \cdot \nabla_{\bx_t} R(\bx_{0|t}, \bc)$, which injects the reward signals into the sampling process:
\begin{equation}
\label{eq:final_guidance_flow}
\begin{aligned}
    \nabla_{\bx_t} R(\bx_{0|t}, \bc) = \nabla_{\bx_t} R(\bx_t - t \cdot {v}_{\theta}(\bx_t,t), \bc) 
    = \frac{\partial R}{\partial \bx_{0\mid t}}\cdot \!\left(\mathbf{I}- t \cdot \frac{\partial {v}_{\theta}(\bx_t,t)}{\partial \bx_t} \right) ,
\end{aligned}
\end{equation}
where the reward function is actually applied on the decoded image domain through the decoder. For simplicity of discussion, we omit the decoder notation. By adding Eq.~\eqref{eq:final_guidance_flow} into Eq.~\eqref{eq:ode_score_guidance}, flow-matching models can also achieve alignment by injecting reward-aware dynamics into the latent states of the next timesteps. Essentially, since modifying intermediate latent states is equivalent to steering the sampling trajectory, we introduce a hypernetwork that absorbs reward guidance. This allows the foundation model to directly output the aligned score as a variation of the original pre-trained model's score, thereby enabling efficient test-time alignment.

\section{More Hypernetwork Details}
\subsection{Architecture of the Hypernetwork}
The hypernetwork generates LoRA weights for all attention projection parameters \textbf{(to.q, to.k, to.v, to.out.0)} of the pre-trained diffusion denoising model, covering $N$ modules in total across all blocks (4 modules for each of all blocks), each with varying input and output dimensionalities. 
$d_{\text{in}}^{i}$ and $d_{\text{out}}^{i}$ denote the input and output dimensionalities of the weight matrix of the $i$-th module. 
The varying dimensionalities introduce difficulties for the hypernetwork when generating LoRA weights for different modules, as fixed-dimensional outputs would simplify the design and training of the hypernetwork. To address this, we follow \cite{Hyperdreambooth} and introduce two learnable auxiliary matrices per module to absorb the dimension variability. We parameterize the LoRA weights of each module as $\bA^{i} \in \mathbb{R}^{d^{i}_{\text{in}} \times r}$ and $\bB^i \in \mathbb{R}^{r \times d^{i}_{\text{out}}}$, which are dynamically produced by hypernetwork for updating the module $i$ through $\Delta \bW^i = \bA^{i} \bB^{i}$. To handle the varying dimensionality, the the Down matrix $\bA^i$ and Up matrix $\bB^i$ of each LoRA adapter are decomposed as $\bA^i = \bA^{i}_{\text{aux}} \bA^{i}_{\text{hyper}}$ with $\bA^{i}_{\text{aux}} \in \mathbb{R}^{d^{i}_{\text{in}} \times a}, \bA^{i}_{\text{hyper}} \in \mathbb{R}^{a \times r}$ and $\bB^{i} = \bB^{i}_{\text{hyper}} \bB^{i}_{\text{aux}}$ with $\bB^{i}_{\text{hyper}} \in \mathbb{R}^{r \times b}, \bB^{i}_{\text{aux}} \in \mathbb{R}^{b \times d^{i}_{\text{out}}}$. Here $r$ is LoRA rank, $r \ll \min(d^i_{\text{in}}, d^i_{\text{out}})$, $a < d^i_{\text{in}}$ and $b < d^i_{\text{out}}$.
The auxiliary matrices $\bA^{i}_{\text{aux}}$ and $\bB^{i}_{\text{aux}}$ are module-specific and learned in the training process, which are not generated by the hypernetwork. $\bA^{i}_{\text{hyper}}$ and $\bB^{i}_{\text{hyper}}$ have unified dimensions $a \times r$ and $r \times b$ and are produced by the hypernetwork. 
In this way, the hypernetwork can generate weights $\bA^{i}_{\text{hyper}}$ and $\bB^{i}_{\text{hyper}}$ with the same dimensionality on the fly and obtain the LoRA weights for all modules with varying dimensions via the combining auxiliary matrices. 

As shown in Fig.~\ref{fig:framework}, the hypernetwork consists of (1) \textbf{{Perception Encoder (Enc)}}: We use downsampling blocks of the pre-trained SD V1.5 U-Net as encoder, owing to its rich priors. It takes standard noisy latent $\bx_t \in \mathbb{R}^{c \times h \times w}$, timestep $t$ and prompt embedding $\bc \in \mathbb{R}^{77 \times 768}$ (CLIP text encoding) as \textbf{input}, and produces features $\bF \in \mathbb{R}^{320\bc \times \frac{h}{8} \times \frac{w}{8}}$. After flattening as $\bF \in  \mathbb{R}^{\frac{hw}{64} \times 320\bc}$, it is projected as context feature, $\bF \in \mathbb{R}^{\frac{hw}{64} \times d_w}$, via a {Linear} layer. $d_w=r(a+b)$ is intermediate dimension; (2) \textbf{Transformer Decoder} consists of $K=4$ standard transformer blocks (containing self-attention, a cross-attention, and a feed forward layer). The head number and dimension of multi-head attention are 8 and $\frac{d_w}{n_\text{head}}$, respectively. It takes zero tokens $\mathbf{0} \in \mathbb{R}^{N \times d_w}$ with sinusoidal positional embeddings and $\bF$ as inputs. The output is passed through a final \textbf{Linear} layer to produce $\Delta\boldsymbol{\theta}_\text{hyper} \in \mathbb{R}^{N \times d_w}$, reshaped as $N$ $\bA^i_\text{hyper} \in \mathbb{R}^{a \times r}$ and $\bB^i_\text{hyper} \in \mathbb{R}^{r \times b}$, $i=1,...,N$. Finally, through combing the auxiliary matrices $\bA^i_\text{aux}, \bB^i_\text{aux}$,$i=1,...,N$, we can obtain the LoRA weights, containing $N$ LoRA with $\bA^i\in \mathbb{R}^{d^{i}_{\text{in}} \times r}$ and $\bB^i \in \mathbb{R}^{r \times d^{i}_{\text{out}}}$, $i=1,...,N$.

\subsection{Detailed Configurations}
Our hypernetwork is optimized with LoRA-modulated denoising trajectories, ensuring train–inference consistency. To ensure training stability, we initialize the auxiliary matrices $\bB^i_\text{aux}$ to zero, so that generated LoRA weights are zero-initialized at the start of training. We provide more details about hypernetwork configurations and model size in Tab.~\ref{tab:configuration}, training setups and cost in Tab.~\ref{tab:train_setups}.

\begin{table*}[t]
\centering
\renewcommand{\arraystretch}{1.0}
\caption{The hypernetwork configuration of HyperAlign.}
\setlength{\tabcolsep}{6.0mm}{
\begin{tabular}{c|cc}
\hline
  & HyperAlign & HyperAlign  \\ 
\hline
Base model & SD V1.5 & FLUX.1.Dev  \\
Hypernetwork size & 0.4 B & 0.6 B  \\
Training parameters  & 0.08 B & 0.2 B  \\
Auxiliary dimension $a$ & 128 & 128  \\
Auxiliary dimension $b$ & 128 & 128  \\
LoRA rank $r$ & 4 & 4  \\
LoRA alpha $\alpha$ & 8 & 8  \\
Number of transformer blocks $K$ & 4 & 4  \\
Number of injected modules $N$ & 128 & 192  \\
\hline
\end{tabular}}
\label{tab:configuration}
\end{table*}

\begin{table*}[t]
\centering
\renewcommand{\arraystretch}{1.0}
\caption{The training details and cost of HyperAlign on SD and FLUX.}
\setlength{\tabcolsep}{5.0mm}{
\begin{tabular}{c|cc}
\hline
  & HyperAlign & HyperAlign  \\ 
\hline
Base model & SD V1.5 & FLUX.1.Dev  \\
Learning rate & 1$e$-5 & 1$e$-5  \\
GradNorm Clipping  & 1.0 & 1.0  \\
Optimizer & AdamW & AdamW  \\
Batch size & 16 & 4  \\
Image size & 512 $\times$ 512 & 720 $\times$ 720  \\
Smapling steps & 50 & 25  \\ \hline
Memory & 70.51 GB & 80.93 GB  \\
GPU-hrs & 8 & 12  \\
\hline
\end{tabular}}
\label{tab:train_setups}
\end{table*}

\begin{table*}[t]
\centering
\renewcommand{\arraystretch}{1.0}
\caption{GenEval Benchmark \cite{geneval} evaluation based on SD V1.5.}
\resizebox{1.0\linewidth}{!}{
\setlength{\tabcolsep}{1.0mm}{
\begin{tabular}{c|ccccccc}
\hline
Method & Overall & Single object & Two object & Counting & Colors & Position  & Color attribution \\ 
\hline
SD V1.5       & 0.39 & 0.95 & 0.39 & 0.30 & 0.73 & 0.04 & 0.05 \\
\hline
$\varepsilon$-greedy       & {0.48} & 0.98 & 0.66 & 0.38 & 0.72 & 0.05 & 0.05 \\
DAS   & 0.34 & 0.89 & 0.31 & 0.22 & 0.63 & 0.04 & 0.04 \\
FreeDoM   & 0.45 & 0.95 & 0.49 & {0.44} & {0.79} & 0.06 & 0.07 \\
DyMO   & 0.50 & 0.98 & 0.60 & {0.55} & {0.79} & 0.05 & {0.14} \\
\hline
AlignProp     & 0.39 & 0.95 & 0.37 & 0.31 & 0.71 & 0.04 & 0.06 \\
Diffusion-DPO         & 0.41 & 0.97 & 0.39 & 0.37 & 0.75 & 0.04 & 0.06 \\
SPO        & 0.40 & 0.96 & 0.36 & 0.34 & 0.73 & 0.05 & 0.06 \\
Diffusion-KTO         & 0.42 & 0.98 & 0.43 & 0.35 & 0.77 & 0.05 & 0.07 \\
DanceGRPO   & 0.41 & 0.95 & 0.46 & 0.32 & 0.72 & {0.07} & 0.07\\
HyperAlign  & {0.52} & {0.98} & {0.66} & 0.40 & {0.82} & {0.09} & {0.24} \\
\hline
\end{tabular}}}
\label{tab:sd_compare_geneval}
\end{table*}

\begin{table*}[ht]
\centering
\renewcommand{\arraystretch}{1.0}
\caption{Comparison results on GenEval \cite{geneval}.}
\resizebox{0.8\linewidth}{!}{
\setlength{\tabcolsep}{1.0mm}{
\begin{tabular}{c|ccccccc}
\hline
Method & Overall & Single object & Two object & Counting & Colors & Position  & Color attribution \\ 
\hline
FLUX        & 0.63 & 0.97 & 0.77 & 0.68 & 0.78 & 0.21 & 0.44 \\
DanceGRPO   & {0.68} & {0.98} & 0.86 & 0.72 & 0.78 & 0.22 & 0.46 \\
\hline
HyperAlign  & {0.70} & {0.98} & {0.88} & 0.70 & {0.88} & 0.20 & {0.54} \\
\hline
\end{tabular}}}
\label{tab:flux_compare_geneval}
\end{table*}

\section{Additional Experimental Details and Results}
\subsection{Additional Qualitative Results}
We provide more visual results for qualitative evaluation. 

\par
\begin{figure}[t]
\centering
 \includegraphics[width=1.0\linewidth]{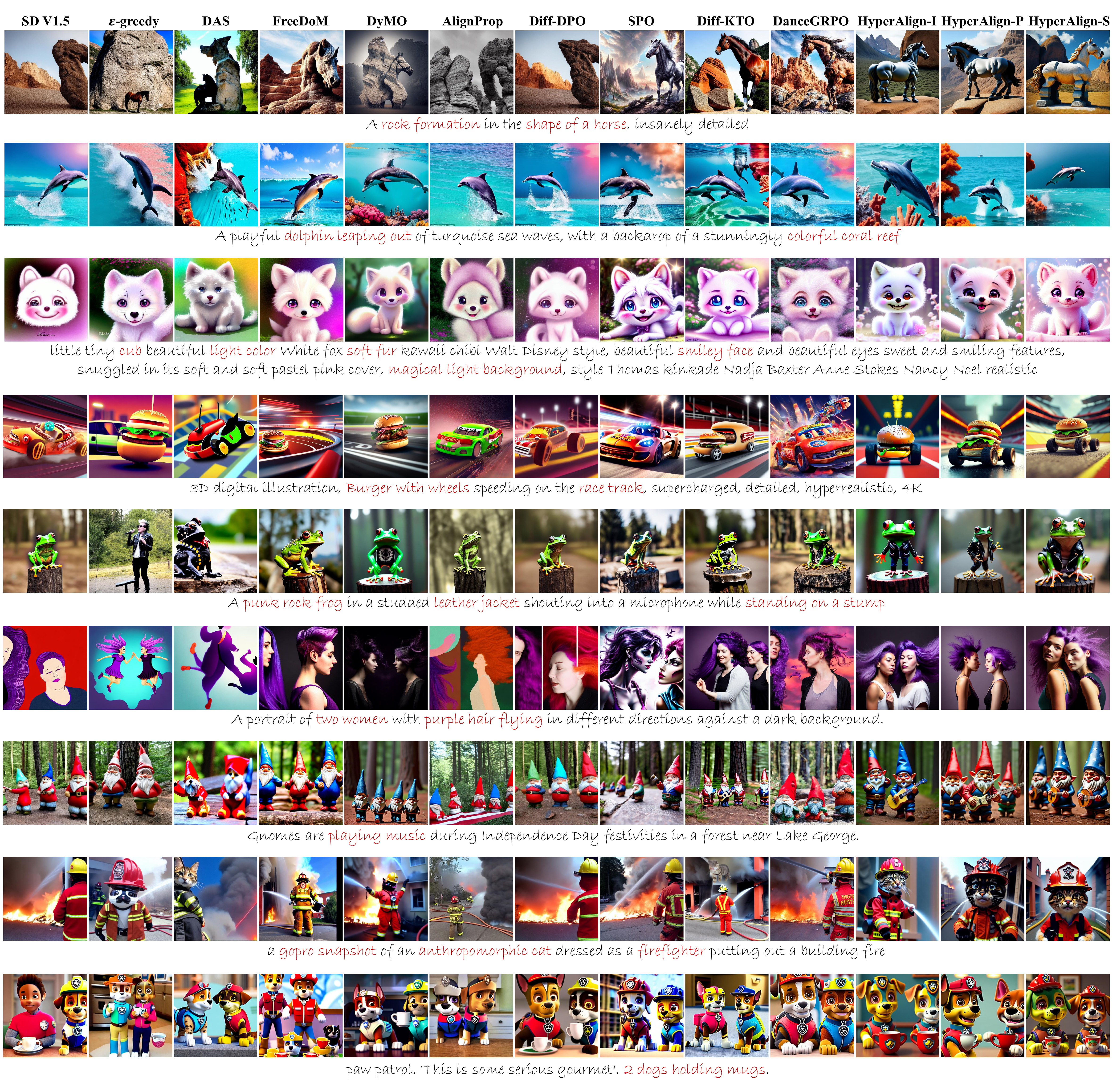}
\caption{Qualitative comparison based on SD V1.5 backbones.}
\label{fig:supply_sd_comparsion}
\end{figure}
\begin{figure}[t]
\centering
 \includegraphics[width=1.0\linewidth]{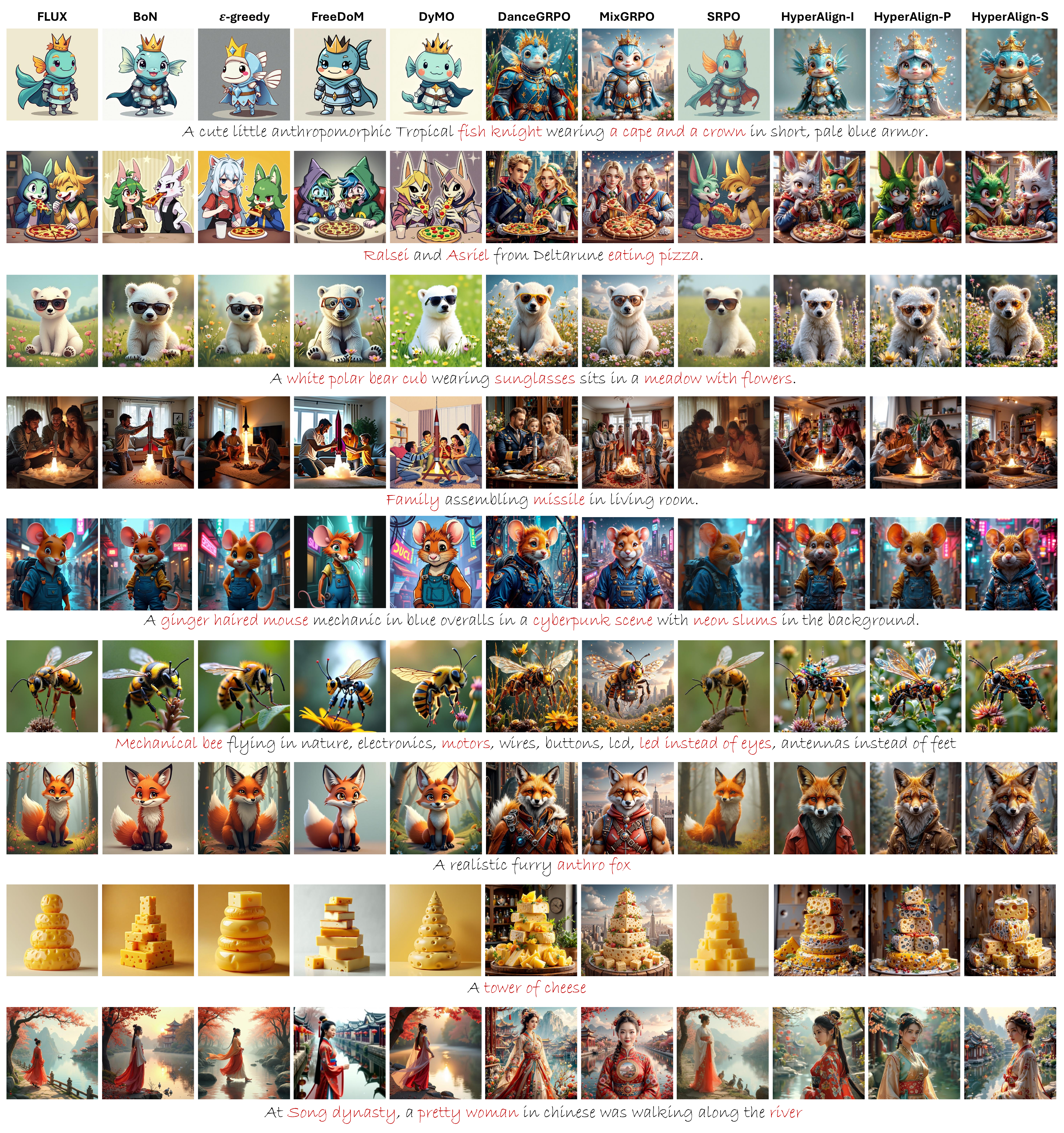}
\caption{Qualitative comparison based on FLUX backbones.}
\label{fig:supply_flux_comparsion}
\end{figure}

\textbf{More results on visual comparison.} 
We provide additional visual comparison results as shown in Fig.~\ref{fig:supply_sd_comparsion} for SD V1.5-based backbones and Fig.~\ref{fig:supply_flux_comparsion} for FLUX-based backbones, respectively.  Our approach generates high-quality images more closely aligned with contextual semantics and better cater to human preferences. All three variants of HyperAlign, differing in the frequency of new LoRA weight generation, achieve strong performance, further demonstrating the effectiveness and flexibility of the proposed framework.

\par
\begin{figure}[t]
\centering
 \includegraphics[width=1.0\linewidth]{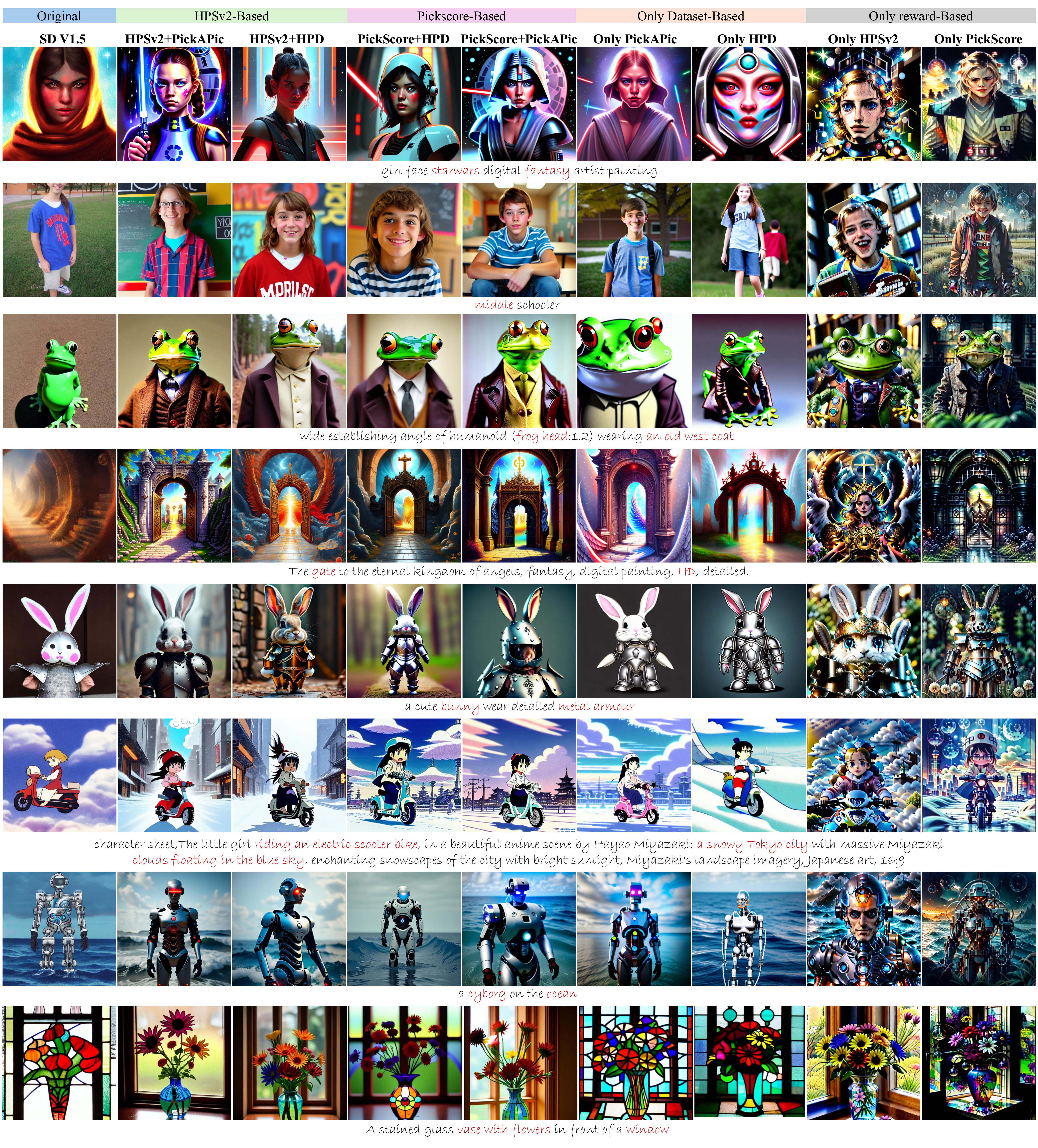}
\caption{The visual results of ablation study.}
\label{fig:visual_ablation}
\end{figure}

\textbf{Qualitative results on ablation studies.} 
In Tab.~\ref{tab:ablation_study}, We report ablation results examining the effects of different reward models and preference datasets. The experimental metrics show that our method remains effective and robust across diverse reward–preference configurations. In Fig.~\ref{fig:visual_ablation}, we visualize the ablation study results. It is observed that the visual qualities of the generated image by our method (HPSv2-based and PickScore-based) are consistent with the numerical results, enhancing both aesthetic appeal and semantic correctness. Compared with SD V1.5 \cite{SD}, supervised fine-tuning solely on preference datasets yields only marginal gains. Additionally, we observe that although reward-only optimization attains higher metric scores, it leads to over-optimized and visually saturated samples, which further demonstrates the effectiveness of our proposed method.

\subsection{Additional Quantitative Results}
\begin{figure}[t]
\centering
 \includegraphics[width=0.8\linewidth]{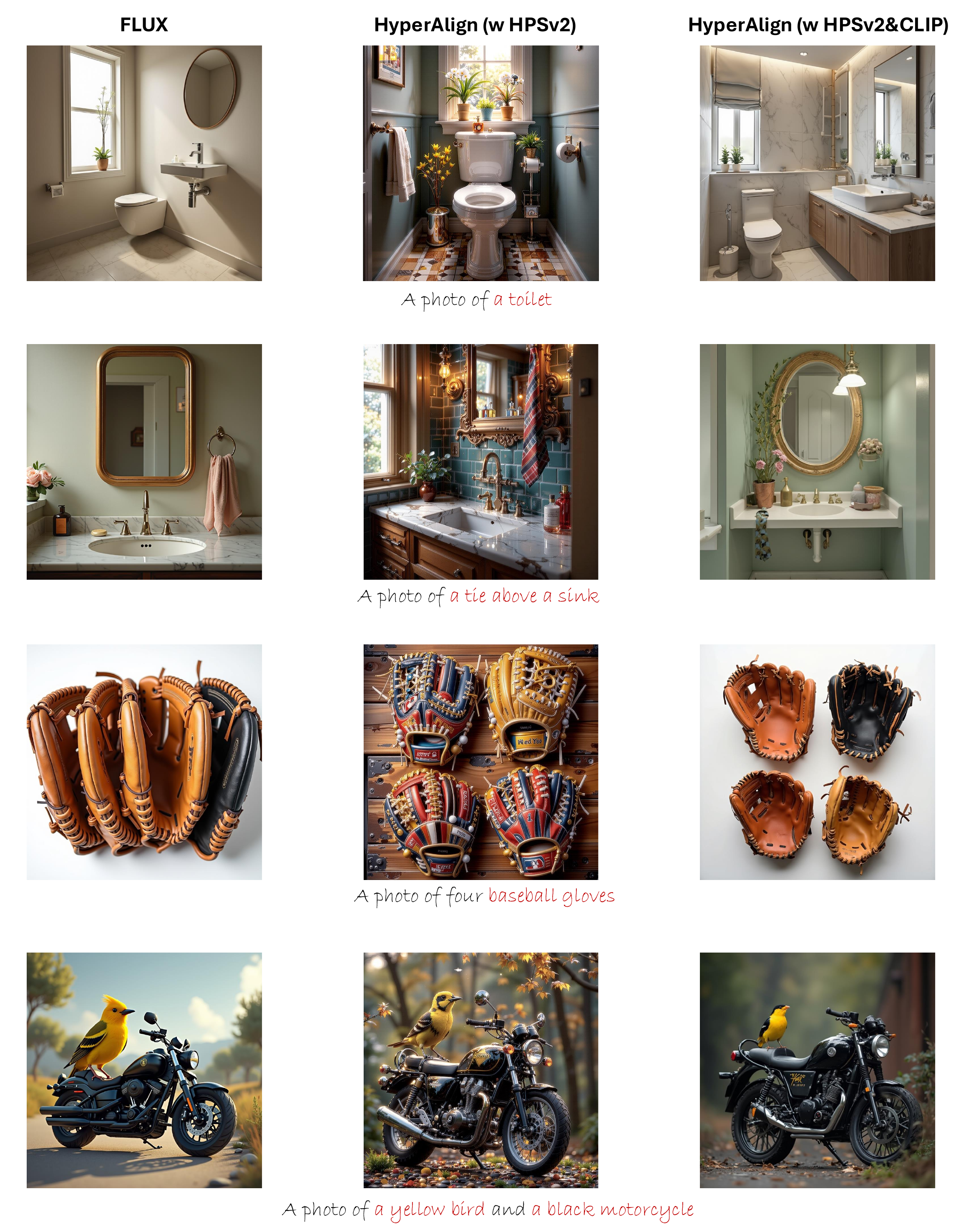}
\caption{The comparison includes the baseline FLUX outputs, the results obtained through HPS-only optimization, and the versions further improved with joint HPS and CLIP objectives.}
\label{fig:supply_flux_align1}
\end{figure}

\begin{figure}[t]
\centering
 \includegraphics[width=0.8\linewidth]{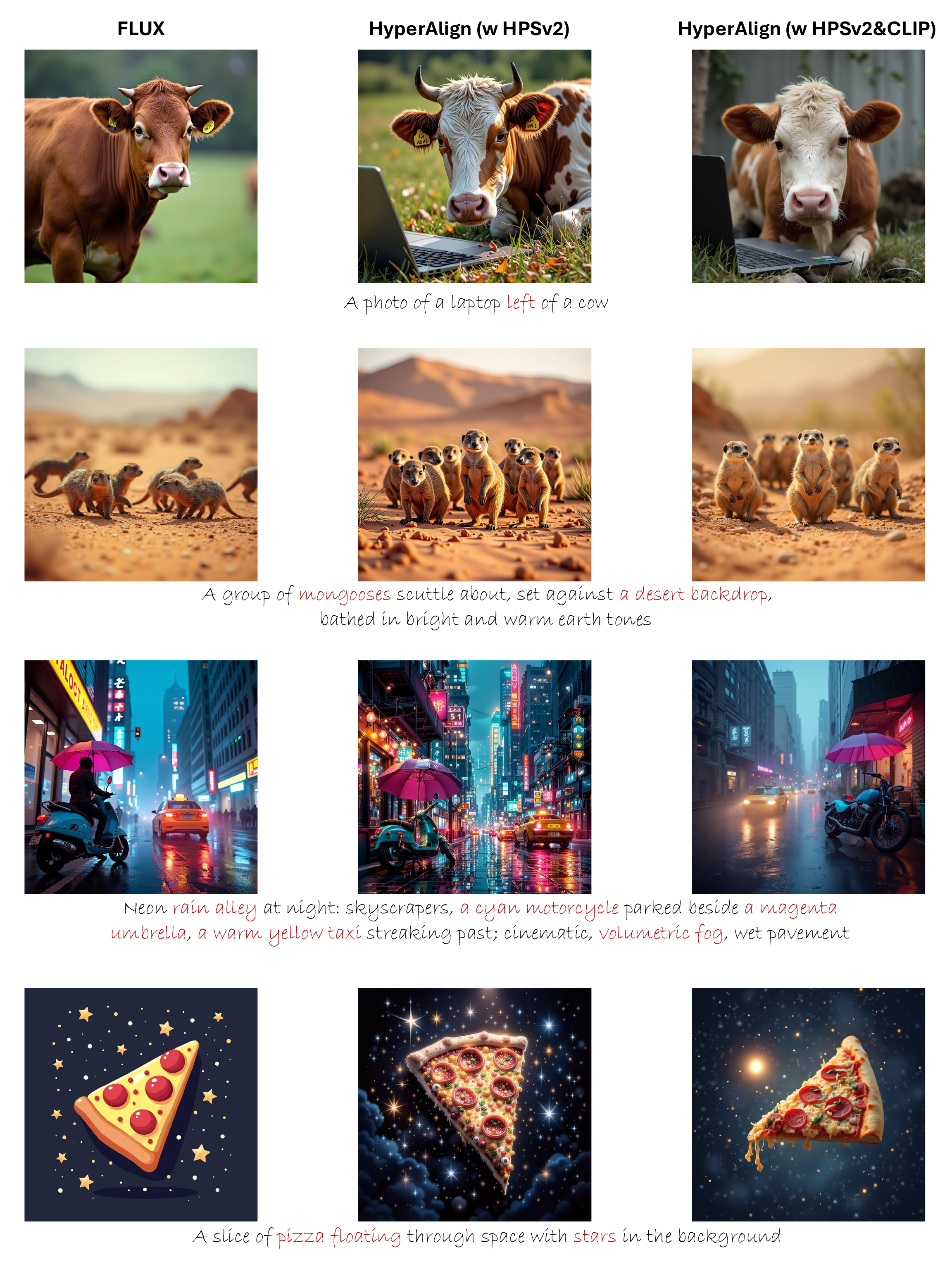}
\caption{The comparison includes the baseline FLUX outputs, the results obtained through HPS-only optimization, and the versions further improved with joint HPS and CLIP objectives.}
\label{fig:supply_flux_align2}
\end{figure}

We conduct quantitative evaluation on GenEval benchmark \cite{geneval} and show comparisons in Tab.~\ref{tab:sd_compare_geneval} for SD V1.5-based backbones. The results show that our method performs very well and shows superiority in many aspects, \eg, overall, attribute binding and object synthesis. To further evaluate the ability to capture high-level semantics, we incorporate the CLIP score into the training objective following DanceGRPO \cite{DanceGRPO}. The main quantitative results on the GenEval benchmark for FLUX-based backbones are presented in Tab.~\ref{tab:flux_compare_geneval}. Corresponding qualitative comparisons are provided in Fig.~\ref{fig:supply_flux_align1} and Fig.~\ref{fig:supply_flux_align2}. Compared with HPS-only optimization, jointly optimizing with both HPS and CLIP objectives yields noticeably better semantic consistency. 

\begin{figure}[t]
    \centering
    \includegraphics[width=1.0\linewidth]{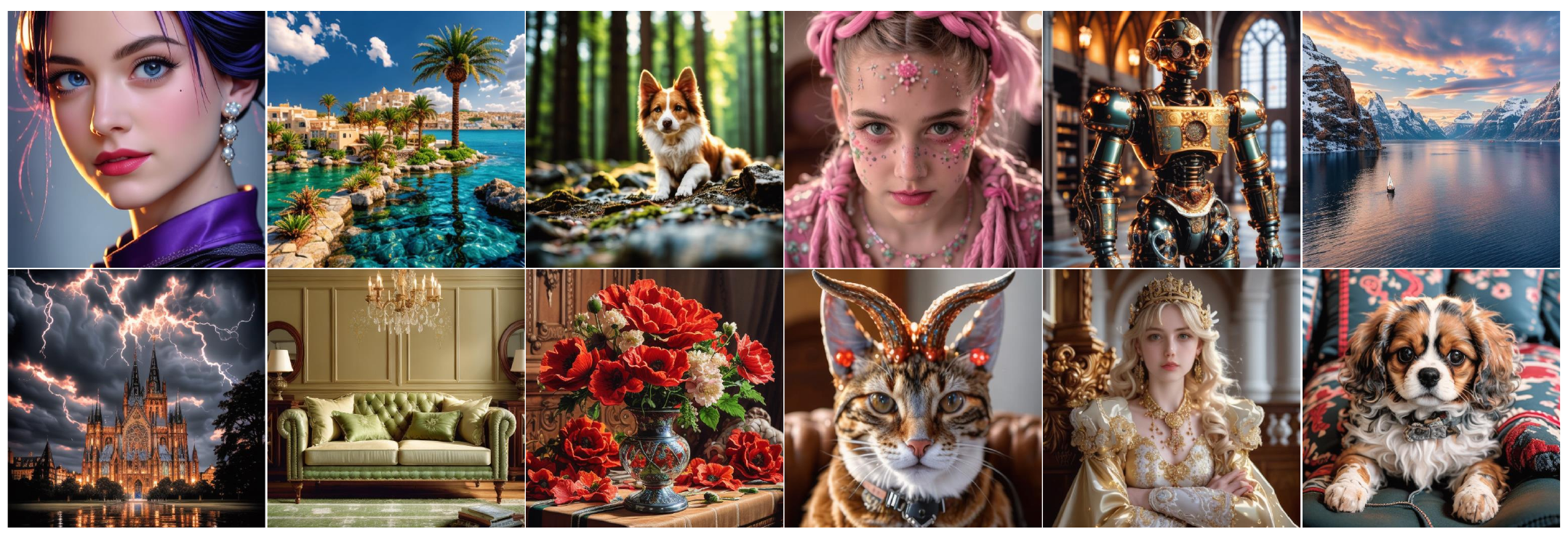}
    \caption{Visual results produced by HyperAlign based on SD-Turbo \cite{SDTurbo} under four-step inference. The used prompts are shown in Tab.~\ref{tab:visual_result_sd_turbo_prompts}.}
    \label{fig:visual_result_sd_turbo_hyperalign}
\end{figure}

\begin{figure}[t]
    \centering
    \includegraphics[width=1.0\linewidth]{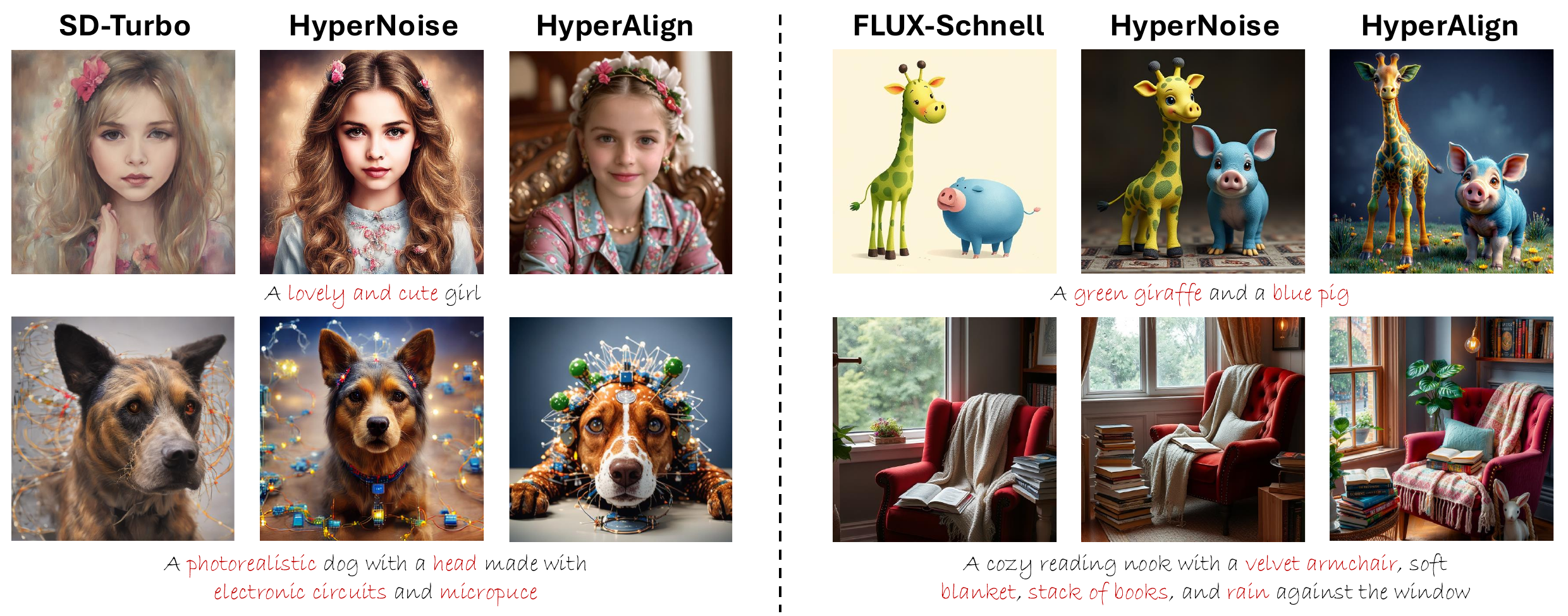}
    \caption{Qualitative comparison with HyperNoise based on SD-Turbo \cite{SDTurbo} and FLUX.1.Schnell \cite{FLUX_schnell}.}
    \label{fig:compare_with_hypernoise}
\end{figure}

\begin{table*}[t]
\centering
\renewcommand{\arraystretch}{1.0}
\caption{Comparison with HyperNoise based on SD-Turbo \cite{SDTurbo} across multiple inference steps.}
\setlength{\tabcolsep}{1.6mm}{
\begin{tabular}{c|ccccccc}
\hline
SD-Turbo & Aes & Pick & IR & CLIP & HPS \\ 
\hline
One-step     & 6.014 & 21.49 & 0.559 & \textbf{0.2685} & 0.2721 \\
+HyperNoise  & \textbf{6.234} & 21.59 & 0.873 & 0.2547 & 0.3106 \\
\rowcolor{tabcolor} +HyperAlign & 6.127 & \textbf{22.06} & \textbf{1.326} & 0.2596 & \textbf{0.3274} \\ \hline
Four-step     & 5.658 & 21.22 & 0.457 & \textbf{0.2592} & 0.2837 \\
+HyperNoise  & 5.745 & 20.96 & 0.776 & 0.2504 & 0.3032 \\
\rowcolor{tabcolor} +HyperAlign & \textbf{6.051} & \textbf{21.59} & \textbf{1.132} & 0.2522 & \textbf{0.3176} \\ \hline
Sixteen-step     & 5.399 & 20.50 & 0.164 & \textbf{0.2499} & 0.2634 \\
+HyperNoise  & 5.473 & 20.56 & 0.255 & 0.2422 & 0.3012 \\
\rowcolor{tabcolor} +HyperAlign & \textbf{5.843} & \textbf{21.11} & \textbf{0.616} & 0.2471 & \textbf{0.3101} \\
\hline
\end{tabular}}
\label{tab:more_compare_with_hypernoise_generalization_sdturbo}
\end{table*}

\begin{table*}[t]
\centering
\renewcommand{\arraystretch}{1.0}
\caption{Comparison with HyperNoise based on FLUX.1.Schnell \cite{FLUX_schnell} across multiple inference steps.}
\setlength{\tabcolsep}{1.6mm}{
\begin{tabular}{c|ccccccc}
\hline
FLUX.1.Schnell & Aes & Pick & IR & CLIP & HPS \\ 
\hline
One-step     & 5.783 & 22.01 & 1.097 & 0.2734 & 0.3016 \\
+HyperNoise  & 5.926 & 22.17 & 1.166 & 0.2752 & 0.3076 \\
\rowcolor{tabcolor} +HyperAlign & \textbf{6.096} & \textbf{22.58} & \textbf{1.333} & \textbf{0.2759} & \textbf{0.3369} \\ \hline
Four-step     & 5.922 & 21.87 & 1.059 & 0.2700 & 0.3003 \\
+HyperNoise  & 5.996 & 22.01 & 1.127 & \textbf{0.2702} & 0.3063 \\
\rowcolor{tabcolor} +HyperAlign & \textbf{6.161} & \textbf{22.13} & \textbf{1.147} & 0.2698 & \textbf{0.3322} \\ \hline
Sixteen-step     & 5.897 & 21.09 & 0.849 & \textbf{0.2643} & 0.2885 \\
+HyperNoise  & 5.912 & 21.51 & 0.852 & 0.2595 & 0.2912 \\
\rowcolor{tabcolor} +HyperAlign & \textbf{6.143} & \textbf{21.61} & \textbf{0.906} & 0.2617 & \textbf{0.3167} \\
\hline
\end{tabular}}
\label{tab:more_compare_with_hypernoise_generalization_flux}
\end{table*}

\subsection{Amortized Efficiency}
Amortizing test-time computation through training to improve inference efficiency is a well-motivated and explored in various ways \cite{HyperNoise,Hyperdreambooth}. 
HyperAlign achieves amortized trajectory alignment by training a hypernetwork to replace expensive inference-time optimization with a single forward pass. 
Although a training process is required (as in all amortization methods), inference efficiency is improved with deployment as a main consideration. 
As reported in Tab.~\ref{tab:train_setups}, HyperAlign requires 8 GPU hours for SD V1.5 and 12 GPU hours for FLUX, substantially less than other training-based methods such as AlignProp (12h) \cite{AlignProp}, DanceGRPO (216h) \cite{DanceGRPO}, and Diffusion-DPO (192h) \cite{DiffusionDPO}. 
While a direct comparison with test-time methods is not entirely equivalent — as training-based amortization trades upfront training cost for reduced per-query inference cost — we evaluate amortized complexity via overall latency  $T_\text{all} = T_\text{train}+T_\text{test} \times P$, where $P$ is the number of inference prompts, to provide a reference point. On the Pick-a-Pic dataset ($P=1000$) with SD V1.5, test-time methods FreeDoM and DyMO require $T_\text{all}=$66h and 83h, respectively. HyperAlign achieves better results with $T_\text{all}=8.83$h, supporting the practical value of amortized alignment for large-scale deployment scenarios.

\subsection{Detailed Comparison with HyperNoise}
Although both HyperNoise \cite{HyperNoise} and HyperAlign leverage hypernetworks for alignment, they operate under fundamentally different modeling. HyperNoise predicts improved initial noise and is designed specifically for one-step distilled generators (for the focus on efficiency and maybe easy training), with post-hoc multi-step inference applied after training. HyperAlign instead predicts input-specific LoRA weights to adjust the denoising trajectory at each step of a general multi-step denoising process. 

Given these differences in scope, a direct comparison is not straightforward; nonetheless, we make efforts toward a fair evaluation on the same base models.
\noindent\textbf{(1) Adapting HyperNoise to multi-step models.}
We attempt to adapt HyperNoise to full-step models, \eg, SD V1.5. However, due to its design of adjusting the initial noise, backpropagating supervision signals across the full multi-step diffusion process causes severe training difficulties, including memory explosion and structural artifacts in generated outputs — a consequence of the supervision backpropagation problem over long denoising chains, where only the initial hyper noise predictor can take the supervision signal. 
\noindent\textbf{(2) Adapting HyperAlign to one-step training of distilled models.}
We also adapt HyperAlign to the one-step training setting used by HyperNoise \cite{HyperNoise}, \ie, distilled models \cite{SDTurbo, FLUX_schnell}. While HyperAlign is originally motivated for multi-step trajectory alignment, the hypernetwork LoRA prediction can be straightforwardly applied to distilled models. We implement HyperAlign on SD-Turbo \cite{SDTurbo} and FLUX.1.Schnell \cite{FLUX_schnell} under the same setting as \cite{HyperNoise}, in a straightforward way, optimized with HPSv2 reward and an auxiliary regularization term for the concentrated supervision credit in \textbf{one-step training}. Numerical results for \textbf{one-step, four-step, and sixteen-step inference} are reported in Tab.~\ref{tab:more_compare_with_hypernoise_generalization_sdturbo} and Tab.~\ref{tab:more_compare_with_hypernoise_generalization_flux} for both HyperAlign and HyperNoise, with the same setting in HyperNoise. 
HyperAlign achieves competitive or superior performance compared to HyperNoise, and consistently improves over baselines as the number of inference steps increases. Additional visual results on SD-Turbo under four-step inference are provided in Fig.~\ref{fig:visual_result_sd_turbo_hyperalign}.

\begin{figure*}[t]
\centering
 \includegraphics[width=0.6\linewidth]{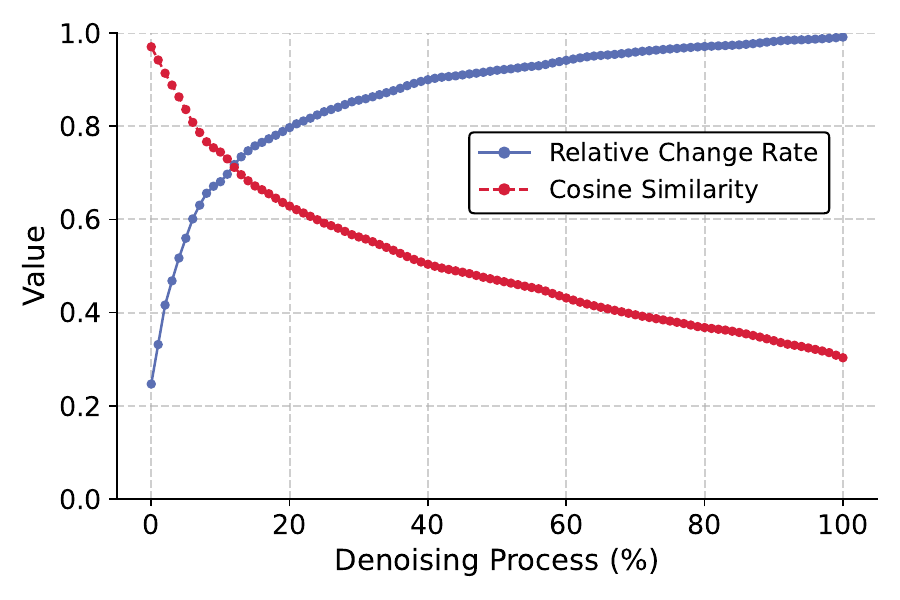}
\caption{Visualization of LoRA weight variations at different timesteps relative to the initial timestep $T$, averaged over 1000 prompts. As the denoising progresses, cosine similarity with the initial weights gradually decreases while the $\ell_1$ relative change rate increases, indicating that LoRA weights evolve significantly across timesteps.
}
\label{fig:lora_change}
\end{figure*}

\begin{figure*}[t]
\centering
 \includegraphics[width=1.0\linewidth]{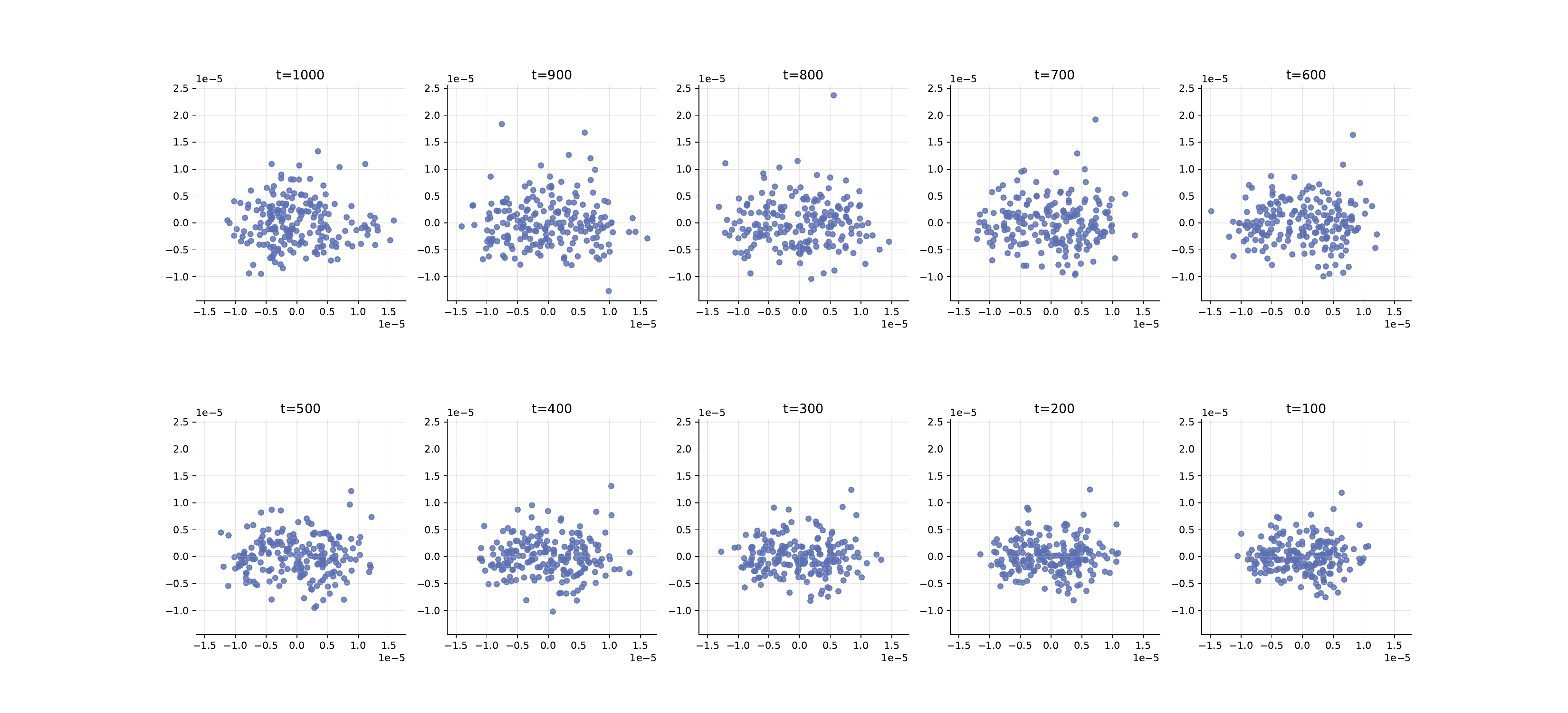}
\caption{Visualization of the statistics of prompt-specific LoRA weights across different steps. The top two PCA components of the LoRAs generated for different prompts (200 examples) at each step are shown. }
\label{fig:weight_var}
\end{figure*}

\subsection{Additional Ablation Study Results}
\par
\textbf{Effect of LoRA rank and transformer decoder depth}. We ablate the LoRA rank ($r$) and the number of transformer decoder blocks ($K$) to measure scalability. As shown in Tab.~\ref{tab:Scalability}, our default configuration is sufficiently expressive and achieves the trade-off between performance and cost.

\par
\textbf{Effect of segments for HyperAlign-P}. Under identical conditions, HyperAlign-I/P/S differ only in LoRA update frequency. HyperAlign-P updates LoRA according to denoising behaviors at different stages shown in Fig.~\ref{fig:curvature}, designed to strike a performance–efficiency balance between HyperAlign-I and HyperAlign-S. To further investigate the effect of segment strategy, we additionally evaluate two HyperAlign variants (named HyperAlign-US3 and HyperAlign-US10), which divide the denoising process uniformly into 3 and 10 segments respectively, updating LoRA once at the start of each segment. As shown in Tab.~\ref{tab:more_ablation_hyper_p}, HyperAlign-P outperform the uniform 3-segment updates and remains competitive with more frequent 10-segment updates, validating our curvature-based segmentation strategy.

\begin{table*}[t]
\centering
\renewcommand{\arraystretch}{1.0}
\caption{Scalability test based on HyperAlign-S.}
\setlength{\tabcolsep}{1.3mm}{
\begin{tabular}{c|c|cccccc}
\hline
Method & Training params  & Aes & Pick & IR & CLIP & HPS  \\ 
\hline
K=4, r=4   &  0.08B  & 5.824 & 22.01 & 0.7731 & 0.2851 & 0.2957 \\
\hline
K=4, r=8   &  0.2B  & 5.835 & 22.06 & 0.7379 & 0.2853 & 0.2913 \\
K=4, r=16   &  0.8B  & 5.871 & 21.99 & 0.7752 & 0.2837 & 0.2962 \\
\hline
K=2, r=4    &  0.05B    & 5.728 & 21.75 & 0.6482 & 0.2833 & 0.2793 \\
K=8, r=4    &  0.12B   & 5.832 & 21.95 & 0.7686 & 0.2851 & 0.2897 \\
\hline
\end{tabular}}
\label{tab:Scalability}
\end{table*}

\begin{table*}[t]
\centering
\renewcommand{\arraystretch}{1.0}
\caption{More ablation study results based on HyperAlign-P.}
\setlength{\tabcolsep}{1.6mm}{
\begin{tabular}{c|ccccccc}
\hline
Method & Aes & Pick & IR & CLIP & HPS \\ 
\hline
HyperAlign-US3     & 5.793 & 21.92 & 0.6107 & 0.2806 & 0.2879 \\
HyperAlign-US10     & 5.868 & 22.00 & 0.6894 & 0.2861 & 0.2899 \\
HyperAlign-P & 5.878 & 21.89 & 0.6979 & 0.2823 & 0.2885 \\
\hline
\end{tabular}}
\label{tab:more_ablation_hyper_p}
\end{table*}

\subsection{Additional Analyses}

\par
\textbf{Evolution of LoRA weights across timesteps.}  
To better understand how generated LoRA weight adapts across the denoising process, we analyze the LoRA weights produced by HyperAlign-S based on SD V1.5 \cite{SD}. As shown in Fig.~\ref{fig:lora_change}, we measure both the cosine similarity and the $\ell_1$ relative change of the LoRA at each timestep relative to the initial timestep $T$. As denoising progresses, cosine similarity steadily decreases while the $\ell_1$ relative change rate consistently increases, indicating that the generated LoRA weights diverge progressively, both in direction and magnitude. This confirms that different denoising stages require functionally distinct adaptations, justifying our design of input-conditioned LoRA generation rather than using a single static LoRA throughout the trajectory.

\par
\textbf{Prompt-specific variation of LoRA weights.}  
To examine whether the hypernetwork generates meaningfully distinct LoRA weights for different inputs, we randomly sample 200 prompts and collect the corresponding generated LoRA parameters at each denoising timestep. We apply PCA to project the high-dimensional LoRA parameters onto their top two principal components for visualization, as shown in Fig.~\ref{fig:weight_var}. The scattered distribution of points confirms that HyperAlign generates input-specific LoRA weights, where different prompts receive distinct adaptations rather than a one-size-fits-all correction.

\subsection{More Details on Human Evaluation}
\label{sec:questionaire}
\begin{figure}[t]
\centering
 \includegraphics[width=0.9\linewidth]{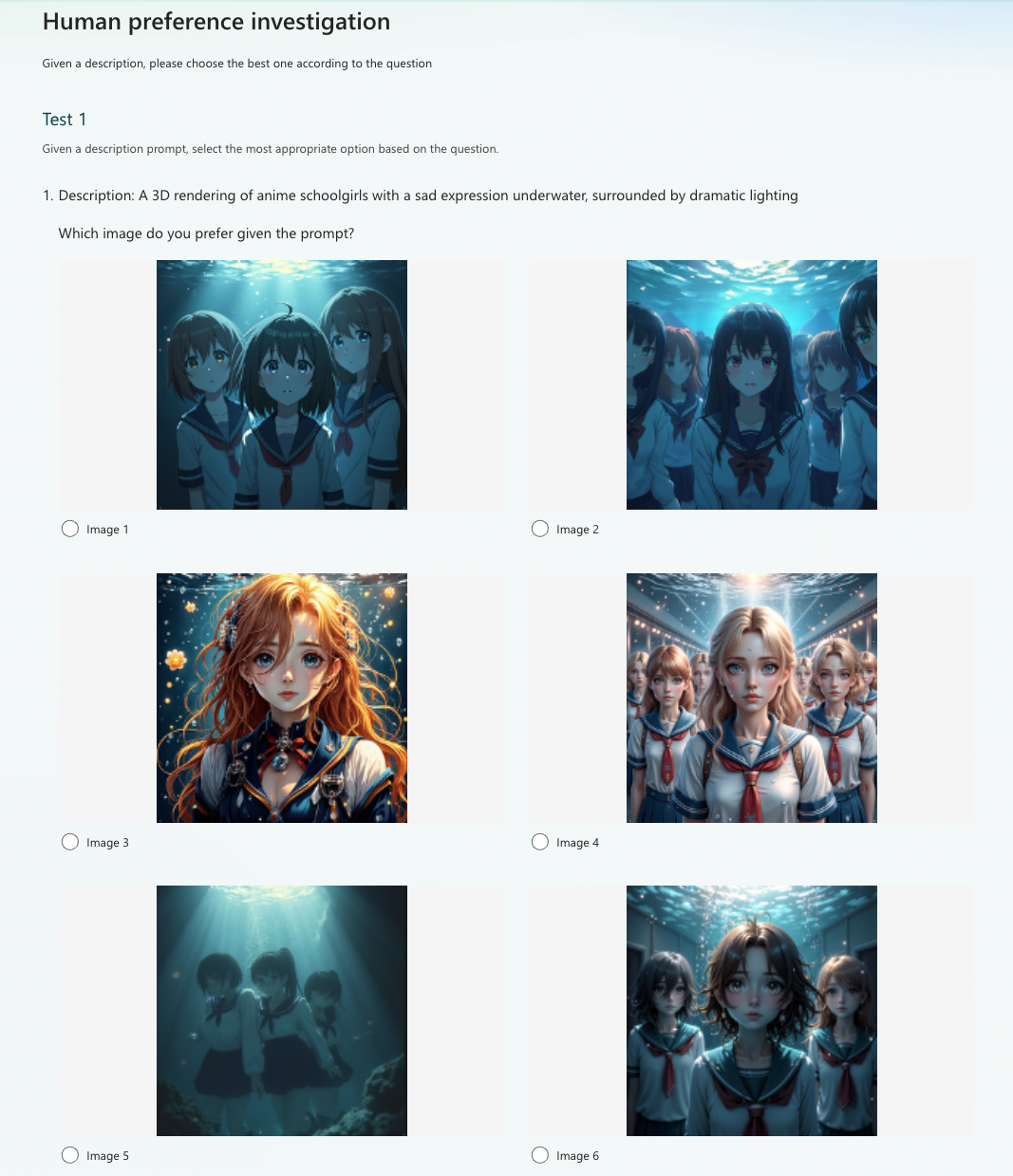}
\caption{The screenshot of human preference investigation: Which image do you prefer given the prompt?}
\label{fig:Q1}
\end{figure}

\begin{figure}[htbp]
\centering
 \includegraphics[width=0.9\linewidth]{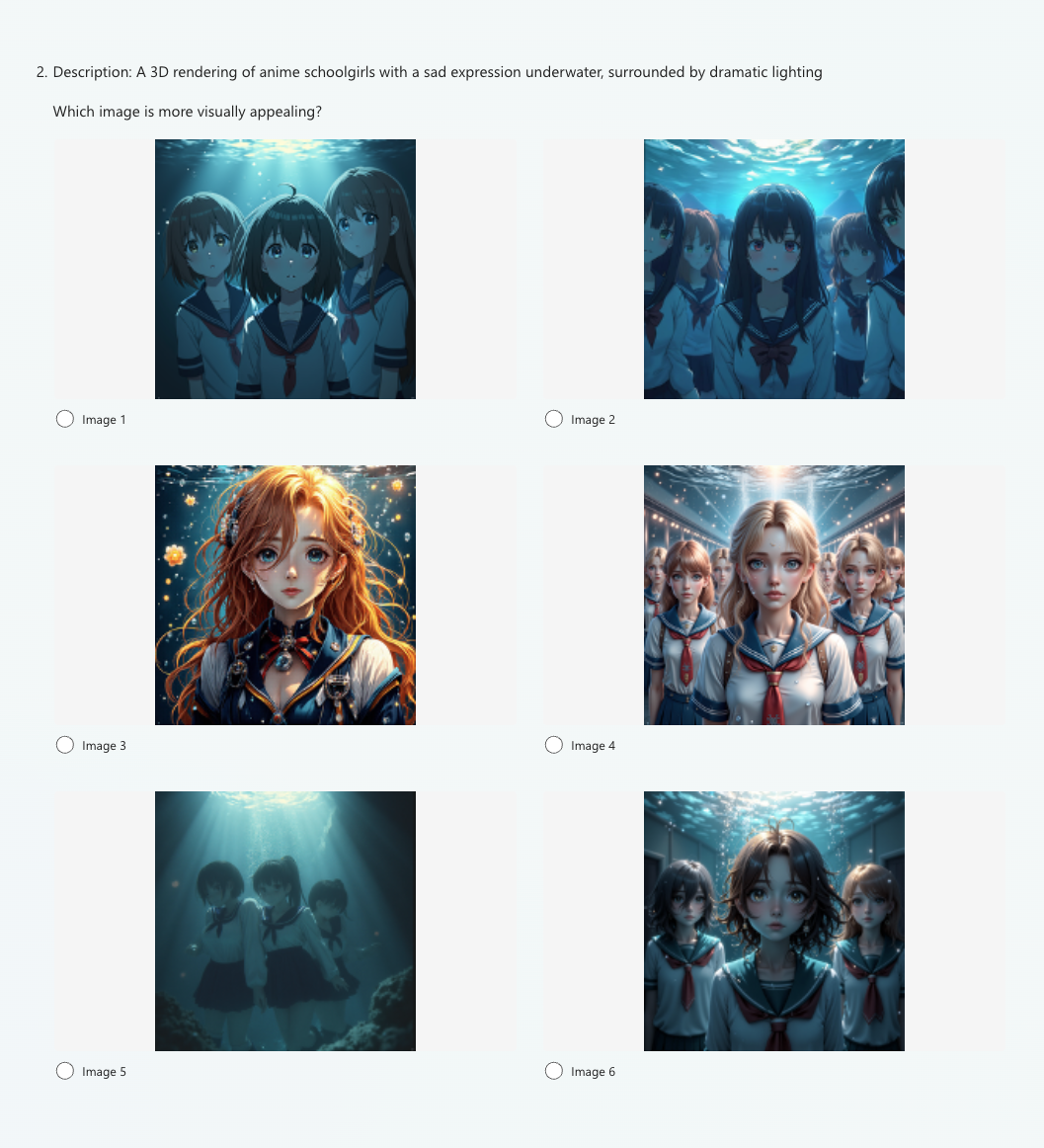}
\caption{The screenshot of human preference investigation: Which image is more visually appealing?}
\label{fig:Q2}
\end{figure}

\begin{figure}[htbp]
\centering
 \includegraphics[width=0.9\linewidth]{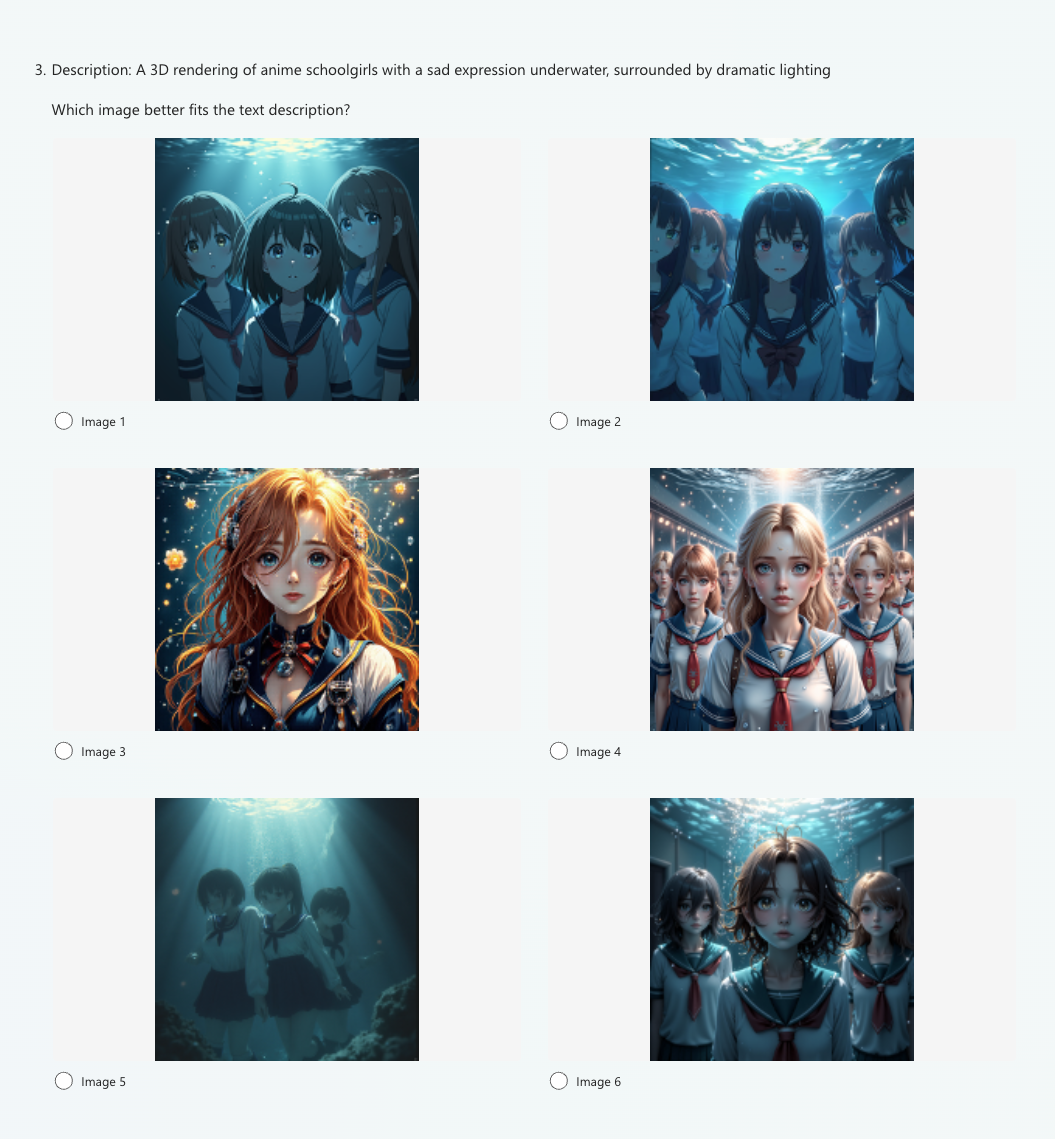}
\caption{The screenshot of human preference investigation: Which image better fits the text description?}
\label{fig:Q3}
\end{figure}

We administer our user study using structured survey forms, where each prompt is presented as an independent section. Within each section, we present multiple images generated by different methods for the same prompt. Participants answer three questions: Q1 (Fig.~\ref{fig:Q1}), Q2 (Fig.~\ref{fig:Q2}) and Q3 (Fig.~\ref{fig:Q3}). Each question targets a different aspect of preference (overall preference, visual appeal, and prompt alignment), and participants are asked to select the most favorable image among the provided options. Participants are recruited via an online platform and remain fully anonymous. To ensure reliable evaluation, all participants are required to hold at least a bachelor’s degree, and their privacy and identity are strictly protected throughout the study.

\section{Ethical and Social Impacts}
HyperAlign is a hypernetwork-based alignment framework for text-to-image diffusion and flow-matching models that enhances semantic consistency and visual quality. As with text-to-image generation broadly, there are general risks associated with this line of research that the community should be aware of. Generative models trained on large-scale web data may reflect societal biases, and alignment methods that optimize for human preferences may inadvertently reinforce such biases if the underlying reward signals or preference datasets lack sufficient diversity. More broadly, advances in controllable image generation increase the potential for misuse, including the creation of misleading, harmful, or privacy-violating content. We encourage the broader research community to invest in content moderation, bias auditing, and responsible deployment practices. 
In this work, the benefits of HyperAlign outweigh these potential concerns. The proposed framework lowers the barrier to high-quality test-time alignment and improves accessibility for diverse user groups. 
HyperAlign is intended to improve alignment quality and accessibility for diverse user groups, and we advocate for its use within transparent and ethically grounded deployment frameworks.

\begin{table*}[t]
\centering
\small
\renewcommand{\arraystretch}{1.0}
\caption{Detailed prompts used for generated images in Fig.~\ref{fig:iteration}.}
\label{tab:fig_iter_prompt_list}
\setlength{\tabcolsep}{1.0mm}{
\begin{tabular}{p{2cm}|p{11cm}}
\hline
\multicolumn{1}{c|}{Image} & \multicolumn{1}{c}{Prompt} \\ \hline

\multicolumn{1}{c|}{\scriptsize{$\bc_1$}} & 
\begin{minipage}[c]{\linewidth}
\vspace*{\fill}
\vspace*{0.5em}
\centering \scriptsize{a gopro snapshot of an anthropomorphic cat dressed as a firefighter putting out a building fire.}
\vspace*{0.5em}
\vspace*{\fill}
\end{minipage} \\ \hline

\multicolumn{1}{c|}{\scriptsize{$\bc_2$}} & 
\begin{minipage}[c]{\linewidth}
\vspace*{\fill}
\vspace*{0.5em}
\centering \scriptsize{pink-haired woman looking straight ahead, full lips, white military clothing with small red details, blue sky with blurred clouds, Chromatic Aberration, Geometric Shape, Photorealistic, Cosmic, Detailed, Bloom, masterpiece, best quality, extremely detailed CG unity 8k wallpaper, landscape, 3D Digital Paintings, award winning photography, Photorealistic, trending on artstation, trending on CGsociety, Intricate, High Detail, dramatic, high quality lighting, vivid anime color.}
\vspace*{0.5em}
\vspace*{\fill}
\end{minipage} \\ \hline

\end{tabular}}
\end{table*}

\begin{table*}[t]
\centering
\small
\renewcommand{\arraystretch}{1.0}
\caption{Detailed prompts used for generated images in Fig.~\ref{fig:visual_result_sd_turbo_hyperalign}.}
\label{tab:visual_result_sd_turbo_prompts}
\setlength{\tabcolsep}{1.0mm}{
\begin{tabular}{p{2cm}|p{9cm}}
\hline
\multicolumn{1}{c|}{Image} & \multicolumn{1}{c}{Prompt} \\ \hline

\multicolumn{1}{c|}{\scriptsize{Row 1, Column 1}} & 
\begin{minipage}[c]{\linewidth}
\vspace*{\fill}
\vspace*{0.5em}
\centering \scriptsize{a goth girl, purple eye shadows, purple lipstick, short black hair, smiling, white teeth, portrait, ultra detailed, octante render, digital art, digital painting, masterpiece, sharp focus, hd, 4k, 8k, hd, high quality, extremely detailed, cinematic lighting, soft illumination, professional shot, award winning, artstation, cgsociety, deviantart.}
\vspace*{0.5em}
\vspace*{\fill}
\end{minipage} \\ \hline

\multicolumn{1}{c|}{\scriptsize{Row 1, Column 2}} & 
\begin{minipage}[c]{\linewidth}
\vspace*{\fill}
\vspace*{0.5em}
\centering \scriptsize{An image of Malta, covered in Palm trees, highly detailed and realistic.}
\vspace*{0.5em}
\vspace*{\fill}
\end{minipage} \\ \hline

\multicolumn{1}{c|}{\scriptsize{Row 1, Column 3}} & 
\begin{minipage}[c]{\linewidth}
\vspace*{\fill}
\vspace*{0.5em}
\centering \scriptsize{tilt-shift photo of a dog in a forest.}
\vspace*{0.5em}
\vspace*{\fill}
\end{minipage} \\ \hline

\multicolumn{1}{c|}{\scriptsize{Row 1, Column 4}} & 
\begin{minipage}[c]{\linewidth}
\vspace*{\fill}
\vspace*{0.5em}
\centering \scriptsize{A girl with pink pigtails and face tattoos.}
\vspace*{0.5em}
\vspace*{\fill}
\end{minipage} \\ \hline

\multicolumn{1}{c|}{\scriptsize{Row 1, Column 5}} & 
\begin{minipage}[c]{\linewidth}
\vspace*{\fill}
\vspace*{0.5em}
\centering \scriptsize{a shiny metallic renaissance steampunk robot in the style of Jan van Eyck.}
\vspace*{0.5em}
\vspace*{\fill}
\end{minipage} \\ \hline

\multicolumn{1}{c|}{\scriptsize{Row 1, Column 6}} & 
\begin{minipage}[c]{\linewidth}
\vspace*{\fill}
\vspace*{0.5em}
\centering \scriptsize{A sail boat entering a majestic fjord landscape in winter.}
\vspace*{0.5em}
\vspace*{\fill}
\end{minipage} \\ \hline

\multicolumn{1}{c|}{\scriptsize{Row 2, Column 1}} & 
\begin{minipage}[c]{\linewidth}
\vspace*{\fill}
\vspace*{0.5em}
\centering \scriptsize{Gothic cathedral in a stormy night.}
\vspace*{0.5em}
\vspace*{\fill}
\end{minipage} \\ \hline

\multicolumn{1}{c|}{\scriptsize{Row 2, Column 2}} & 
\begin{minipage}[c]{\linewidth}
\vspace*{\fill}
\vspace*{0.5em}
\centering \scriptsize{a leather pistachio-green Italian sofa, pillows on the sofa, cream colored wall.}
\vspace*{0.5em}
\vspace*{\fill}
\end{minipage} \\ \hline

\multicolumn{1}{c|}{\scriptsize{Row 2, Column 3}} & 
\begin{minipage}[c]{\linewidth}
\vspace*{\fill}
\vspace*{0.5em}
\centering \scriptsize{Still life, poppies in the vase, vine, bread, natural moody lighting, close up shot, dramatic lighting, early morning backlight, high details, 32k, by Brothers Hildebrandt, by Peder Monsted, maximum detail, in the style of richard schmid, Jeremy Mann, Daniel F Gerhartz, Aaron Grffin ultra detailed.}
\vspace*{0.5em}
\vspace*{\fill}
\end{minipage} \\ \hline

\multicolumn{1}{c|}{\scriptsize{Row 2, Column 4}} & 
\begin{minipage}[c]{\linewidth}
\vspace*{\fill}
\vspace*{0.5em}
\centering \scriptsize{A cat with two horns on its head.}
\vspace*{0.5em}
\vspace*{\fill}
\end{minipage} \\ \hline

\multicolumn{1}{c|}{\scriptsize{Row 2, Column 5}} & 
\begin{minipage}[c]{\linewidth}
\vspace*{\fill}
\vspace*{0.5em}
\centering \scriptsize{Portrait of an anime princess in white and golden clothes.}
\vspace*{0.5em}
\vspace*{\fill}
\end{minipage} \\ \hline

\multicolumn{1}{c|}{\scriptsize{Row 2, Column 6}} & 
\begin{minipage}[c]{\linewidth}
\vspace*{\fill}
\vspace*{0.5em}
\centering \scriptsize{blue merle toy aussie cavalier king charles mixed breed.}
\vspace*{0.5em}
\vspace*{\fill}
\end{minipage} \\ \hline

\end{tabular}}
\end{table*}

\end{document}